\documentclass[]{fairmeta}
\usepackage[table]{xcolor}
\usepackage{amssymb}
\usepackage{amsmath}
\usepackage{dsfont}
\usepackage{algorithm}
\usepackage{algpseudocode}

\usepackage{caption}
\usepackage{xcolor}
\usepackage{graphicx}
\usepackage{xparse}
\usepackage{enumitem}
\usepackage{pdfpages}
\usepackage{titletoc}
\usepackage[notransparent]{svg}
\NewDocumentCommand{\inlineimage}{O{0.5} m}{%
  \raisebox{-0.2\baselineskip}{\includegraphics[height=#1\baselineskip]{#2}}\hspace{-3pt}
}

\definecolor{verylightblue}{HTML}{f4fbff}
\definecolor{lightblue}{HTML}{77C9FF}
\definecolor{mediumblue}{HTML}{0099FF}

\definecolor{verylightred}{HTML}{fffbf4}
\definecolor{lightred}{HTML}{FF5957}
\definecolor{mediumred}{HTML}{ff002b}

\tcbset{
  aibox/.style={
    width=480pt,
    top=7pt,
    bottom=5pt,
    colback=metabg,
    colframe=black,
    colbacktitle=black,
    enhanced,
    center,
    attach boxed title to top left={yshift=-0.1in,xshift=0.15in},
    boxed title style={boxrule=0pt,colframe=white,},
  }
}
\newtcolorbox{AIbox}[2][]{aibox,title=#2,#1}

\usepackage{xcolor} %
\usepackage{tikz}
\usetikzlibrary{fit,calc}
\newcommand*{\tikzmk}[1]{\tikz[remember picture,overlay,] \node (#1) {};\ignorespaces}

\newcommand{\boxit}[1]{\tikz[remember picture,overlay]{\node[xshift=-7.5em,yshift=-1em,fill=#1,opacity=.25,fit={(A)($(B)+(1.1\linewidth,1\baselineskip)$)}] {};}\ignorespaces}

\newcommand{\boxittwo}[1]{\tikz[remember picture,overlay]{\node[xshift=-4.15em,yshift=-1em,fill=#1,opacity=.25,fit={(A)($(B)+(1.025\linewidth,1\baselineskip)$)}] {};}\ignorespaces}

\newcommand{\boxithree}[1]{\tikz[remember picture,overlay]{\node[xshift=-4.15em,yshift=-1em,fill=#1,opacity=.25,fit={(A)($(B)+(0.475\linewidth,1\baselineskip)$)}] {};}\ignorespaces}

\colorlet{mypink}{red!30}
\colorlet{myblue}{orange!30}
\colorlet{mypurple}{green!10}

\newcommand{\modelname}{\texttt{SPICE}}
\newcommand{\C}{\texttt{Challenger}}
\newcommand{\R}{\texttt{Reasoner}}
\newcommand{\longname}{Self-Play In Corpus Environments}

\definecolor{customblue}{HTML}{0099ff}

\definecolor{customred}{HTML}{ff006b}

\newcommand{\logo}{\raisebox{.15\height}{\inlineimage[0.65]{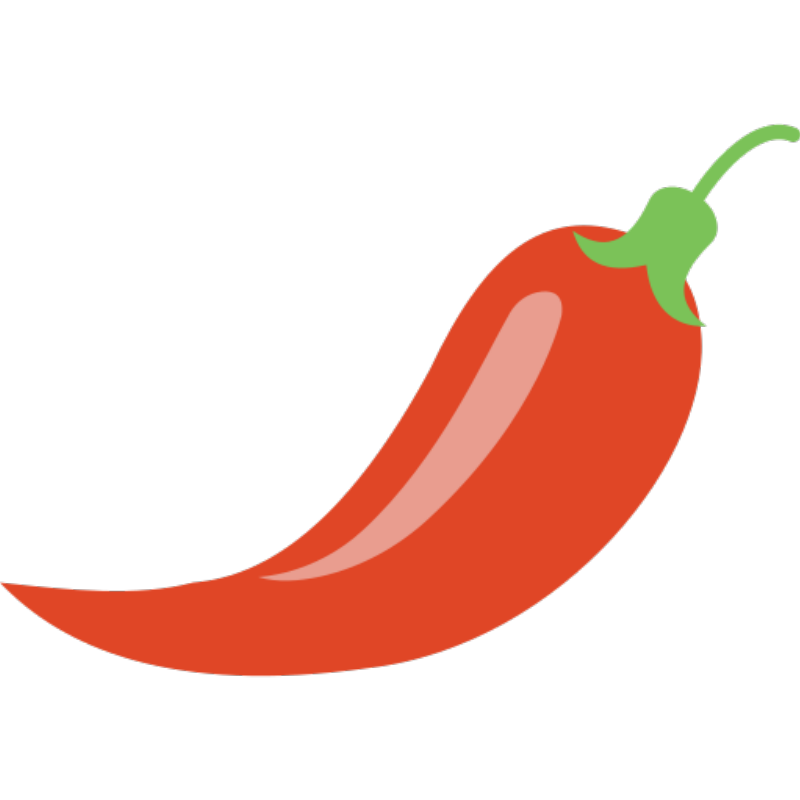}}}

\title{\textcolor{red!90!black}{SPICE} \logo: \textcolor{red!90!black}{S}elf-\textcolor{red!90!black}{P}lay \textcolor{red!90!black}{I}n \textcolor{red!90!black}{C}orpus%
~\textcolor{red!90!black}{E}nvironments Improves Reasoning}

\affiliation[1]{FAIR at Meta}
\affiliation[2]{National University of Singapore}

\contribution[*]{Work done at Meta}
\contribution[\ddagger]{Joint second author}
\contribution[\nabla]{Joint last author}

\author[1,2,*]{Bo Liu}
\author[1,\ddagger]{Chuanyang Jin}
\author[1,\ddagger]{Seungone Kim}
\author[1]{Weizhe Yuan}
\author[1]{Wenting Zhao}
\author[1]{Ilia Kulikov}
\author[1]{Xian Li}
\author[1]{Sainbayar Sukhbaatar}
\author[1,\nabla]{Jack Lanchantin}
\author[1,\nabla]{Jason Weston}

\abstract{
Self-improving systems require environmental interaction for continuous adaptation. We introduce \modelname{} (\longname{}), a reinforcement learning framework where a single model acts in two roles: a \C{} that mines documents from a large corpus to generate diverse reasoning tasks, and a \R{} that solves them. Through adversarial dynamics, the \C{} creates an automatic curriculum at the frontier of the \R{}'s capability, while corpus grounding provides the rich, near-inexhaustible external signal necessary for sustained improvement. Unlike existing ungrounded self-play methods that offer more limited benefits, \modelname{} achieves consistent gains across mathematical (+8.9\%) and general reasoning (+9.8\%) benchmarks on multiple model families. Our analysis reveals how document grounding is a key ingredient in \modelname{} to continuously generate its own increasingly challenging goals and achieve them, enabling sustained self-improvement.
}

\date{\today}
\correspondence{
\email{\textcolor{red}{benjaminliu.eecs@gmail.com}}, \email{\textcolor{red!90!black}{\{jacklanchantin, jase\}@meta.com}}
}

\begin{document}

\maketitle

\begin{figure}[th]
    \centering    \includegraphics[width=1\linewidth]{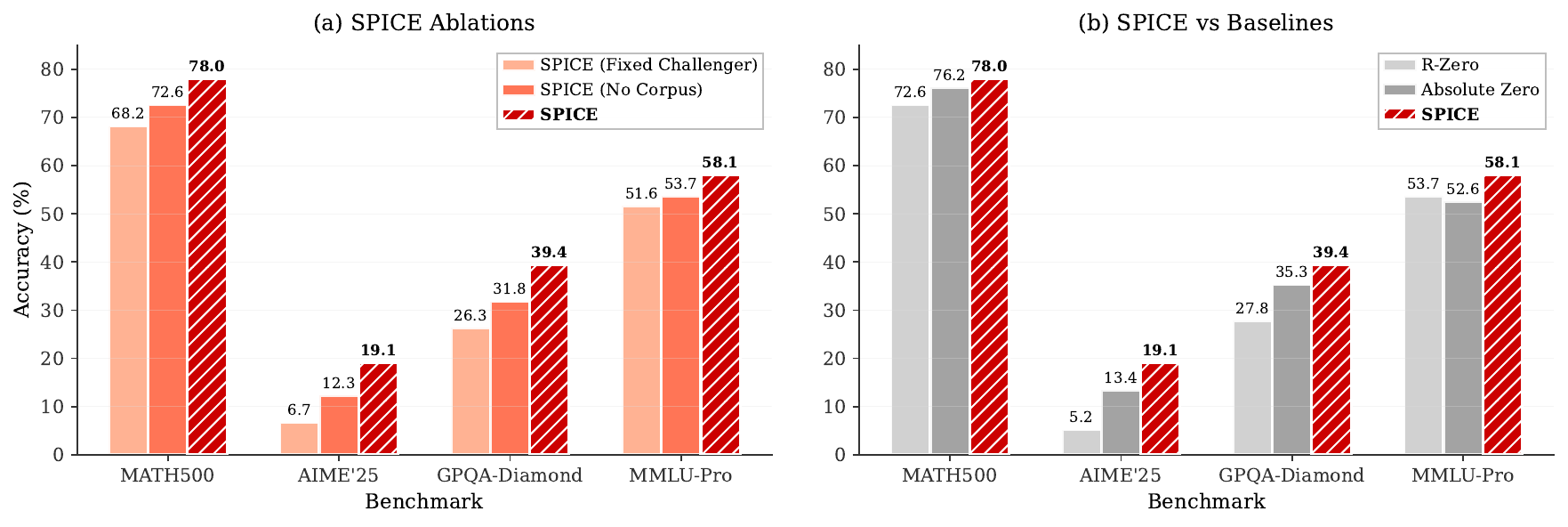}
    
    \caption{\modelname{} (\longname{}) outperforms state-of-the-art self-play methods for LLMs on Qwen3-4B-Base (right). Training the model in \modelname{} to be both a \C{} and \R{}, creating and solving challenging {\em corpus-grounded} tasks for itself via self-play RL is crucial to this success (left), see ablations for details.
    }
    \vspace{-2pt}
    \label{fig:bar_comparison}
\end{figure}

\section{Introduction}
\label{section:intro}

Self-improving artificial intelligence~\citep{schmidhuber2007godel,clune2019ai} has long been envisioned as a path toward artificial general intelligence, where systems autonomously enhance their capabilities through environmental interaction and continuous adaptation. This vision is becoming tangible with large language models (LLMs;~\cite{kaplan2020scaling, achiam2023gpt,dubey2024llama}), which demonstrate remarkable reasoning abilities across diverse domains~\citep{wei2022chain,kojima2022large}. Recent breakthroughs show that reinforcement learning can unlock more sophisticated reasoning. Models like OpenAI o1~\citep{openai2024o1} and DeepSeek-R1~\citep{deepseek2024r1} achieve expert-level performance on mathematical and coding tasks through reinforcement learning with verifiable rewards (RLVR). To scale these capabilities without human supervision, self-play offers a promising paradigm~\citep{silver2017mastering,sukhbaatar2018intrinsicmotivationautomaticcurricula}, where models improve by competing against themselves and generating automatic feedback through competition.

However, most existing self-play methods for language models achieve initial improvements but quickly face fundamental barriers. Without external grounding, models inevitably plateau or collapse~\citep{huang2025r,chen2025self,kuba2025language} due to two critical issues: (1) hallucination amplification, where factual errors in both generated questions and answers compound as models train on their own unverifiable synthetic data, and (2) information symmetry, where both the problem generator and solver share the same knowledge base, preventing genuine challenge and leading to simpler, more repetitive patterns. Even approaches maintaining diversity through variational synthesis~\citep{liang2025beyond} ultimately remain bounded by their initial coverage, which is merely a compressed representation of the original pretraining data~\citep{morris2025much}. These systematic empirical failures indicate that self-improvement requires interaction with an external source providing diverse, verifiable feedback, rather than closed-loop pure introspection.

We introduce \modelname{} (\longname{}), a self-play reinforcement learning framework that mines contexts from a large document corpus to generate diverse reasoning tasks with document-grounded answers. A single model acts in two roles: a \C{} that constructs a curriculum of such challenging document-grounded tasks, and a \R{} that develops robust reasoning capabilities by solving the tasks without document access.  
A key component is information asymmetry: the \C{} grounds questions and gold answers in retrieved documents unseen by the \R{}, creating genuine challenge. The vast diversity of documents ensures continual novelty beyond the model's internalized knowledge. Simultaneously, corpus grounding prevents hallucination by anchoring both questions and gold answers in real-world content rather than model-generated fantasies, ensuring factual accuracy throughout the self-play loop. 

During self-play, the \C{} is rewarded for generating problems at the frontier of the \R{}'s capability (maximizing variance in success rates), while the \R{} is rewarded for correct answers. This adversarial yet symbiotic interaction with corpus grounding enables the system to continuously discover new challenges grounded in real knowledge and overcome them. Importantly, our approach relies on raw documents without predefined questions or labels. Tasks are generated with diverse formats (multiple-choice questions and free-form questions with integer/expression/string answers) which serve as universal verifiers, enabling self-play across any language domain without requiring specialized executors or rule-based validators. This breaks the verification bottleneck that has confined prior work to narrow domains like mathematics and code, while document-grounded answers ensure verification remains factually anchored.

Training with corpus-grounded self-play produces reasoning capabilities that transfer broadly across diverse model choices. On Qwen3-4B-Base, \modelname{} achieves 44.9\% average performance versus 35.8\% baseline performance (+9.1\% absolute gain), while Qwen3-8B-Base improves from 43.0\% to 48.7\% (+5.7\%). OctoThinker-3B-Hybrid-Base shows the largest gains from 14.7\% to 25.2\% (+10.5\%), and OctoThinker-8B-Hybrid-Base improves from 20.5\% to 32.4\% (+11.9\%), where \modelname{} surpasses both standard RLVR and pure (ungrounded) self-play baselines in all cases. These gains span both mathematical reasoning (average +8.9\%) and general reasoning tasks (+9.8\% across MMLU-Pro, GPQA-Diamond, SuperGPQA, and BBEH), demonstrating that corpus grounding develops broadly applicable capabilities. The adversarial dynamics between \C{} and \R{} create an automatic curriculum: the fixed \R{}'s pass rate decreases from 55\% to 35\% as it learns to generate progressively harder problems, while the fixed \C{}'s pass rate increases from 55\% to 85\%, indicating successful co-evolution of both roles. Corpus grounding proves critical for sustained improvement: without it, models show limited improvement and fail to maintain the adaptive challenge necessary for sustained learning. With corpus grounding, the system maintains stable advancement throughout training by continuously mining new document contexts for novel challenges.

Overall, our work makes the following contributions:
\begin{enumerate}
    \item We show that treating a large document corpus as an external knowledge source enables sustained self-improvement, outperforming ungrounded methods that rely solely on the model's intrinsic knowledge.
    \item We propose corpus-grounded self-play where a model acting in two roles (\C{} and \R{}) generates tasks with document-extracted answers and solves them, using diverse task formats that enable verification across all domains without specialized tools.
    \item We empirically validate that corpus-grounded self-play achieves consistent improvements across mathematical and general reasoning tasks, overcoming the domain-specific limitations of existing approaches.
\end{enumerate}

\modelname{} presents a paradigm shift in self-improving reasoning methods: 
from closed-loop self-play that often stagnates due to hallucination drift, to open-ended improvement through interaction with the vast, verifiable knowledge embedded in web document corpora.

\section{SPICE: Self-Play In Corpus Environments}
\label{sec:method}

\begin{figure}
    \centering
    \includegraphics[width=1\linewidth]{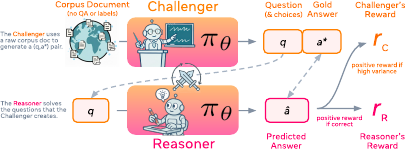} 
    \vspace{-1.5em}
    \caption{\modelname{} is a self-play framework where a single LLM, $\pi_\theta$, acts in two roles: a \C{} ($\text{role}=C$), which poses difficult questions, and a \R{} ($\text{role}=R$), which tries to correctly answer such questions. The Challenger uses a raw document (which does not contain existing questions or labels) from a corpus to generate a ($q$, $a^*$) pair. Depending on the contents of the document, the \C{} can generate either a closed-form multiple choice question or a free-form question with different answer types (integer, float, expression). The \C{} gets rewarded for questions with high variance in \R{}'s correctness, and the \R{} gets rewarded for answering questions correctly.}
    \label{fig:spice_method}
    \vspace{-0.5em}
\end{figure}

\modelname{} is an end-to-end framework featuring a single model that acts in two roles through self-play: a \C{} (C) and \R{} (R). When playing the role of \C{}, the model grounds itself in web documents to propose questions that challenge the \R{}. Subsequently, the model switches roles to the \R{} to answer the questions. This iterative process, governed by adversarial dynamics, allows both roles to co-evolve, leading to a progressively more capable model, with the corpus providing the necessary external signal for sustained improvement. The entire framework is self-supervised, requiring no human intervention, only a large unstructured corpus. We overview \modelname{} in \autoref{fig:spice_method}, and detail the training method in \autoref{alg:spice-training}. In the following subsections, we outline the different roles during self-play.

\subsection{Notations and Preliminaries}
\label{sec:prelim}
We formulate corpus-grounded self-play as a game where a single model $\pi_\theta$ acts in two roles: generating questions from documents ($\text{role}=C$) or answering questions ($\text{role}=R$). Let $\mathcal{D}$ denote a corpus of documents, $\mathcal{Q}$ the space of questions, and $\mathcal{A}$ the space of answers. When $\text{role}=C$, the model has access to document $d \sim \mathcal{D}$; when $\text{role}=R$, it only sees question $q \in \mathcal{Q}$, creating information asymmetry between the roles.

\subsection{Challenger: Document-Grounded Task Generation}
When acting as \C{} (C), the model $\pi_\theta$ learns to generate problems that maximally challenge the \R{} while remaining solvable. Given a corpus $\mathcal{D}$, the model in $\text{role}=C$ produces diverse tasks as follows.

\textbf{Document Sampling.} We uniformly sample passages from a large document corpus, extracting segments as documents $d \in \mathcal{D}$. Each document provides context for generating questions with verifiable answers extracted directly from the text.

\textbf{Multi-Format Task Generation.} Given document $d$, the \C{} takes multiple attempts to create valid question-answer pairs. The model generates question $q$ and extracts gold answer $a^*$ directly from the document:
$$(q, a^*) \sim \pi_\theta(\cdot | d, \text{role}=C)$$

The \C{} selects between two formats based on document evaluation (see Appendix~\ref{app:prompt_templates}): (i) multiple-choice questions (MCQ) with four options and a document-grounded correct answer, or (ii) free-form questions with typed answers (integer, expression, string) extracted from the document. A generated question $q$ is considered \textit{valid} if it is formatted correctly and parsable from the generation. Our prompt (details given in Appendix~\ref{app:prompt_templates}) guides the \C{} through multi-step complex information extraction, difficulty enhancement, and self-testing to ensure questions are challenging yet solvable without the source document.

\textbf{Variance-Based Curriculum Reward.} For each valid question $(q, a^*)$, we sample $K$ responses $\hat{a}_i$ from the \R{} and compute the \C{}'s reward using a scaled variance of the responses:
\begin{equation}
\label{eq:challenger_reward}
r_C(q,a^*) = \begin{cases}
\exp\left(-\frac{(\text{Var}(\{l_1, ..., l_K\}) - 0.25)^2}{2 \cdot 0.01}\right) & \text{if } q \text{ is valid} \\
\rho & \text{otherwise (penalty)}
\end{cases}
\end{equation}
where $l_i = \mathds{1}[\hat{a}_i = a^*]$. This Gaussian-shaped reward function is scaled to [0,1], maximizing at 1.0 when variance equals 0.25 (50\% pass rate), indicating optimal task difficulty. Tasks that are too easy or too hard receive exponentially lower rewards; as the \R{} improves, the \C{} is rewarded for increasing task difficulty, creating an automatic curriculum.

\subsection{Reasoner: Solving Tasks Without Document Access}

When acting as \R{} (R), the model $\pi_\theta$ learns to solve the \C{}'s tasks without document access.

\textbf{Answer Generation.} Given valid question $q$ alone, the model generates answer $\hat{a} \sim \pi_\theta(\cdot | q, \text{role}=R)$. The model is prompted to reason step by step and place its final answer within $\texttt{\textbackslash boxed\{\}}$ tags, thus forcing it to rely solely on internalized knowledge.

\textbf{Binary Correctness Reward.} The \R{} receives a correctness reward $r_R(\hat{a}, a^*) = \mathds{1}[\hat{a} = a^*]$ conditioned on a rule-based verifier that checks answer equivalence with the document-extracted gold answer.

\subsection{Training with Role-Specific Advantages}

We optimize both roles jointly (shared weights) by maximizing expected rewards:
\begin{equation}
J(\theta) = \mathbb{E}_{d \sim \mathcal{D}} \left[ \mathbb{E}_{(q,a^*) \sim \pi_\theta(\cdot|d, \text{role}=C)} [r_C(q,a^*)] + \mathbb{E}_{\hat{a} \sim \pi_\theta(\cdot|q, \text{role}=R)} [r_R(\hat{a}, a^*)] \right]
\end{equation}

We use DrGRPO~\citep{liu2025understanding} with a separate advantage computation for each role. Given \C{} trajectories with variance rewards and \R{} trajectories with correctness rewards, we compute role-specific advantages:
\vspace{-0.5em}
\begin{align}
\hat{A}_C^i &= r_C^i - \text{mean}(\{r_C^j\}_j) \\
\hat{A}_R^i &= r_R^i - \text{mean}(\{r_R^j\}_j)
\end{align}

By centering returns around role-specific expectations without standard deviation normalization, we ensure gradient updates reflect genuine learning signals rather than difficulty-induced noise.

\subsection{Implementation}

To implement \modelname{}, we develop a self-play reinforcement learning system for finetuning LLMs. Our training framework builds on Oat~\citep{liu2024oat}, which provides interfaces for a distributed actor-learner architecture~\citep{espeholt2018impala}. We instantiate actors to execute the self-play loop, using vLLM~\citep{kwon2023efficient} for efficient model inference during both \C{} and \R{} generation phases. 

The actors generate experiences by alternating between roles: first generating questions with document access (as \C{}), then answering them without document access (as \R{}). The resulting trajectories are sent to the collocated learner, which updates the LLM via DrGRPO with role-specific advantages. The responses sampled from the \R{} serve dual purposes: computing variance for the \C{}'s reward and forming the trajectory group for the \R{}'s update. We implement answer verification using Math-Verify\footnote{\url{https://github.com/huggingface/Math-Verify}}, which handles equivalence checking for mathematical expressions and exact matching for other answer types. The complete training procedure is detailed in Algorithm~\ref{alg:spice-training}.

\begin{algorithm}[t]
\caption{\textcolor{red!90!black}{\modelname{}} \raisebox{-0.2em}{\includegraphics[height=1.2em]{figs/SPICE_logo.pdf}}: Self-Play In Corpus Environments}
\label{alg:spice-training}
\begin{algorithmic}[1]
\Require Pretrained LLM $\pi_\theta$; corpus $\mathcal{D}$; batch size $B$; group size $G$; iterations $T$; penalty $\rho$
\For{$t \gets 1$ to $T$}
  \tikzmk{A}\State \textbf{Challenger Role:} Generate challenging problems with document access
  \For{$b \gets 1$ to $B$}
    \State Sample document $d \sim \mathcal{D}$
    \State Generate multiple attempts: $\{(q_i, a^*_i)\}_{i=1}^{N} \gets \pi_\theta(d, \text{role}=C)$ \Comment{MCQ or free-form}
    \State $\mathcal{T}_C \gets$ Subsample $G$ trajectories preserving valid:invalid ratio
    \If{$q_i \in \mathcal{T}_C$ is valid}
      \State $\{\hat{a}_k\}_{k=1}^{G} \gets \pi_\theta(q_{i}, \text{role}=R)$ \Comment{No document access}
      \State $r_C(q_i, a^*_i) \gets$ Compute reward using Eq.~\eqref{eq:challenger_reward} \Comment{Gaussian variance reward}
    \Else
      \State $r_C(q_i, a^*_i) \gets \rho$ \Comment{Invalid task penalty}
    \EndIf
  \EndFor
  \tikzmk{B}
  \boxit{myblue}
  \tikzmk{A}\State \textbf{Reasoner Role:} Solve generated problems without document access
  \State Select a random valid task $(q, a^*)$ from Challenger phase
  \State $\mathcal{T}_R \gets \{\hat{a}_i\}_{i=1}^{G} \sim \pi_\theta(q, \text{role}=R)$ \Comment{$G$ responses for training}
  \For{$i \gets 1$ to $G$}
    \State $r_R(\hat{a}_i, a^*_i) \gets \mathds{1}[\hat{a}_i = a^*]$ \Comment{Binary correctness reward}
  \EndFor
  \tikzmk{B}
  \boxittwo{mypink}
  \tikzmk{A} \State \textbf{Update Phase:} Optimize $\pi_\theta$ with role-specific advantages
  \For{trajectory $i$ in $\mathcal{T}_C$}
    \State $\hat{A}_C^i \gets r_C^i - \text{mean}(\{r_C^j : j \in \mathcal{T}_C\})$ \Comment{DrGRPO}
  \EndFor
  \For{trajectory $i$ in $\mathcal{T}_R$}
    \State $\hat{A}_R^i \gets r_R^i - \text{mean}(\{r_R^j : j \in \mathcal{T}_R\})$ \Comment{DrGRPO}
  \EndFor
  
  \State Update $\pi_\theta$ via policy gradient with advantages $\{\hat{A}_C^i\}$ and $\{\hat{A}_R^i\}$
  \tikzmk{B}
  \boxithree{mypurple}
\EndFor
\State \Return Trained model $\pi_\theta$
\end{algorithmic}
\end{algorithm}

\section{Experimental Results}
\label{sec:experiments}

\subsection{Setup}
\label{sec:main_setup}

\textbf{Training Configuration.} Following Algorithm~\ref{alg:spice-training}, we train on corpus $\mathcal{D}$ with $|\mathcal{D}| = 20,000$ documents from high-quality, freely available sources. For mathematics, we use Nemotron-CC-Math~\citep{mahabadi2025nemotron}; for general reasoning, we use documents from NaturalReasoning~\citep{yuan2025naturalreasoning}, a subset of DCLM~\citep{li2024datacomp}. We extract document segments of up to 5,992 tokens to fit within the model's context window. Each iteration attempts to generate at least one valid task, ($q, a^*$), per document with up to $N=1024$ attempts. While multiple valid questions may be generated, we randomly select one per document to balance the training data size between \C{} and \R{} roles. We use temperature 1.0 for both roles, with group size $G=8$ responses per question serving dual purposes: computing variance for the \C{}'s reward and forming the group for DrGRPO advantage computation. The penalty for invalid questions is set to $\rho = -0.1$. We train for a fixed $T=640$ iterations with batch size $B=128$. Detailed training configurations for all methods are provided in Appendix~\ref{app:training_details}.

\textbf{Models and Baselines.} We evaluate \modelname{} on four base models: Qwen3-4B-Base, Qwen3-8B-Base \citep{yang2025qwen3}, OctoThinker-3B-Hybrid-Base, and OctoThinker-8B-Hybrid-Base \citep{wang2025octothinker}. We compare against several baselines: (1) \textbf{Base Model}, the pretrained model without any post-training, establishing the starting performance; (2) \textbf{Strong Challenger}, which uses a fixed stronger model (Qwen3-32B-Instruct) as the \C{} to generate questions with our model training only as \R{}, testing whether a stronger question generator improves learning; (3) \textbf{R-Zero}~\citep{huang2025r}, pure self-play without corpus grounding where the model generates its own questions from scratch without document access, representing ungrounded self-play; and (4) \textbf{Absolute Zero}~\citep{zhao2025absolute}, self-play restricted to code generation tasks with Python execution as verification, representing domain-specific grounded self-play. We run all baseline methods ourselves using their publicly available code to ensure fair comparison with identical training infrastructure, unified evaluation protocols, and model-specific prompt templates detailed in Appendix~\ref{app:prompt_templates}, with training configurations provided in Appendix~\ref{app:training_details}.

\textbf{Evaluation Benchmarks.} We conduct a comprehensive evaluation across mathematical and general reasoning tasks. For mathematical reasoning, we evaluate on MATH-500~\citep{hendrycksmath2021}, a curated subset of challenging competition problems; OlympiadBench~\citep{he2024olympiadbench}, featuring olympiad-level problems; Minerva Math~\citep{lewkowycz2022solving}, covering STEM topics; GSM8K~\citep{cobbe2021gsm8k}, grade-school word problems; and AMC~\citep{amc}, AIME'24, AIME'25~\citep{aime} competition problems from the American Mathematics Competitions. For general reasoning, we use SuperGPQA~\citep{du2025supergpqa}, a large-scale benchmark targeting graduate-level reasoning across 285 disciplines with Google-search-resistant questions; GPQA-Diamond~\citep{rein2023gpqagraduatelevelgoogleproofqa}, graduate-level questions resistant to pattern-matching; MMLU-Pro~\citep{wang2024mmlupro}, a challenging multi-task understanding dataset; and BBEH~\citep{bbeh}, extending BIG-Bench Hard with more complex reasoning tasks. We use the \texttt{simple-evals}\footnote{\url{https://github.com/openai/simple-evals}} framework with GPT-4o~\citep{openai2024gpt4ocard} for answer equivalence checking. Most evaluations use greedy decoding with training-consistent prompts following~\cite{ma2025general}, except AIME'24 and AIME'25 which are averaged over 32 sampling runs following~\cite{simplerl}. Detailed evaluation settings are provided in Appendix~\ref{app:eval_settings}.

\begin{table*}[t]
\centering
\small
\caption{Comprehensive evaluation across mathematical and general reasoning benchmarks comparing \modelname{} against state-of-the-art self-play methods for LLMs. All methods use self-play except ``Strong Challenger'', which uses the fixed Qwen-32B-Instruct model for question generation. Best results per base model are in \textbf{bold}. \modelname{} consistently outperforms all baselines across all four model types.
}
\vspace{-0.1in}
\label{tab:main_results}
\resizebox{\textwidth}{!}{%
\setlength{\tabcolsep}{4pt}
\begin{tabular}{@{}l>{\columncolor{gray!10}}c|ccccccc|cccc|c@{}}
\toprule
& & \multicolumn{7}{c|}{\textbf{Mathematical Reasoning}} & \multicolumn{4}{c|}{\textbf{General Reasoning}} & \\
\cmidrule(lr){3-9} \cmidrule(lr){10-13}
~&
~&
~&
~&
\textbf{MATH}&
~&
~&
\textbf{AIME}&
\textbf{AIME}&
\textbf{Super-} &
\textbf{GPQA-} &
\textbf{MMLU-} & & 
\\
\textbf{Method} & \textbf{Overall} & \textbf{AMC} & \textbf{Minerva} & \textbf{500} & \textbf{GSM8K} & \textbf{Olymp.} & \textbf{24} & \textbf{25} & \textbf{GPQA} & \textbf{Diamond} & \textbf{Pro} & \textbf{BBEH} & \textbf{$\Delta$ vs Base} \\
\midrule
\multicolumn{14}{@{}l}{\cellcolor{green!4}\textit{Qwen3-4B-Base}} \\
Base Model & 35.8 & 47.5 & 42.3 & 68.2 & 72.6 & 34.8 & 10.3 & 6.7 & 25.4 & 26.3 & 51.6 & 8.1 & -- \\
\quad + Strong Challenger & 43.0 & 57.5 & 44.9 & 78.0 & 91.6 & 41.5 & 12.7 & 12.9 & 27.4 & 37.9 & 56.1 & 12.3 & +7.2 \\
\quad + R-Zero & 39.5 & 45.0 & 52.2 & 72.6 & 90.5 & 39.7 & 10.1 & 5.2 & 27.8 & 27.8 & 53.7 & 10.4 & +3.7 \\
\quad + Absolute Zero & 40.7 & 50.0 & 41.9 & 76.2 & 89.3 & 41.5 & 12.2 & 13.4 & 27.1 & 35.3 & 52.6 & 8.3 & +4.9 \\
\rowcolor{red!10}
\quad + \textbf{SPICE (ours)} & \textbf{44.9} & \textbf{57.5} & \textbf{51.9} & \textbf{78.0} & \textbf{92.7} & \textbf{42.7} & \textbf{12.2} & \textbf{19.1} & \textbf{30.2} & \textbf{39.4} & \textbf{58.1} & \textbf{12.3} & \textbf{+9.1} \\
\midrule
\multicolumn{14}{@{}l}{\cellcolor{green!4}\textit{Qwen3-8B-Base}} \\
Base Model & 43.0 & 57.5 & 49.3 & 74.4 & 91.2 & 40.4 & 15.3 & 12.1 & 31.0 & 33.3 & 58.1 & 10.5 & -- \\
\quad + Strong Challenger & 45.6 & 60.0 & 52.5 & 78.2 & 92.4 & 41.6 & 15.3 & 12.1 & 35.0 & 35.8 & 64.8 & 14.4 & +2.6 \\
\quad + R-Zero & 46.3 & 70.0 & 59.2 & 78.0 & 91.8 & 41.3 & 11.7 & 14.2 & 32.8 & 36.4 & 61.7 & 12.0 & +3.3 \\
\quad + Absolute Zero & 46.5 & 62.5 & 52.9 & 76.6 & 92.0 & 47.8 & 18.4 & 18.2 & 33.5 & 36.8 & 62.5 & 10.8 & +3.5 \\
\rowcolor{red!10}
\quad + \textbf{SPICE (ours)} & \textbf{48.7} & \textbf{70.0} & \textbf{59.2} & \textbf{79.4} & \textbf{92.7} & \textbf{42.5} & \textbf{18.4} & \textbf{18.2} & \textbf{35.7} & \textbf{39.4} & \textbf{65.0} & \textbf{14.9} & \textbf{+5.7} \\
\midrule\midrule
\multicolumn{14}{@{}l}{\cellcolor{orange!5}\textit{OctoThinker-3B-Hybrid-Base}} \\
Base Model & 14.7 & 15.0 & 14.3 & 38.2 & 44.3 & 10.8 & 3.4 & 3.5 & 12.8 & 3.5 & 14.9 & 1.0 & -- \\
\quad + Strong Challenger & 21.0 & 15.0 & 15.1 & 39.7 & 73.7 & 14.4 & 0.3 & 0.1 & 15.7 & 25.3 & 24.2 & 7.0 & +6.3 \\
\quad + R-Zero & 20.3 & 20.4 & 22.1 & 48.0 & 74.5 & 15.4 & 0.2 & 0.4 & 12.6 & 6.6 & 18.7 & 4.0 & +5.6 \\
\quad + Absolute Zero & 21.7 & 17.5 & 18.7 & 43.2 & 75.8 & 13.2 & 2.7 & 0.5 & 17.7 & 19.2 & 24.3 & 6.1 & +7.0 \\
\rowcolor{red!10}
\quad + \textbf{SPICE (ours)} & \textbf{25.2} & \textbf{22.5} & \textbf{22.1} & \textbf{48.0} & \textbf{78.8} & \textbf{15.4} & \textbf{3.4} & \textbf{3.5} & \textbf{18.2} & \textbf{26.8} & \textbf{28.9} & \textbf{9.3} & \textbf{+10.5} \\
\midrule
\multicolumn{14}{@{}l}{\cellcolor{orange!5}\textit{OctoThinker-8B-Hybrid-Base}} \\
Base Model & 20.5 & 27.5 & 22.1 & 44.2 & 68.6 & 16.7 & 2.7 & 0.8 & 11.4 & 15.7 & 14.7 & 0.6 & -- \\
\quad + Strong Challenger & 28.2 & 25.0 & 27.2 & 54.4 & 86.4 & 20.0 & 6.6 & 3.3 & 17.6 & 27.8 & 32.9 & 9.5 & +7.7 \\
\quad + R-Zero & 29.9 & 32.5 & 33.1 & 58.4 & 85.2 & 22.6 & 9.9 & 1.9 & 17.9 & 21.7 & 37.4 & 7.8 & +9.4 \\
\quad + Absolute Zero & 29.4 & 32.5 & 34.9 & 56.8 & 87.0 & 25.6 & 0.1 & 3.3 & 18.8 & 27.8 & 31.4 & 5.0 & +8.9 \\
\rowcolor{red!10}
\quad + \textbf{SPICE (ours)} & \textbf{32.4} & \textbf{35.0} & \textbf{34.9} & \textbf{58.4} & \textbf{87.3} & \textbf{25.6} & \textbf{9.9} & \textbf{7.4} & \textbf{18.8} & \textbf{31.8} & \textbf{37.4} & \textbf{9.5} & \textbf{+11.9} \\
\bottomrule
\end{tabular}
}
\vspace{-0.05in}
\end{table*}

\subsection{Quantitative Analysis}

\autoref{tab:main_results} shows the main results on the four different base models we use. \modelname{} consistently outperforms the baselines across all four model families, delivering the largest overall improvements versus the base models: +9.1 (Qwen3‑4B-Base), +5.7 (Qwen3‑8B-Base), +10.5 (OctoThinker‑3B-Hybrid-Base), and +11.9 (OctoThinker‑8B-Hybrid-Base). We observe improvements across \textit{both} math and general reasoning tasks, highlighting the benefits of using a diverse corpus of documents.

\begin{figure}[htbp]
    \centering
    
    \begin{minipage}[b]{0.48\textwidth}
    \centering
    \includegraphics[width=1.0\linewidth]{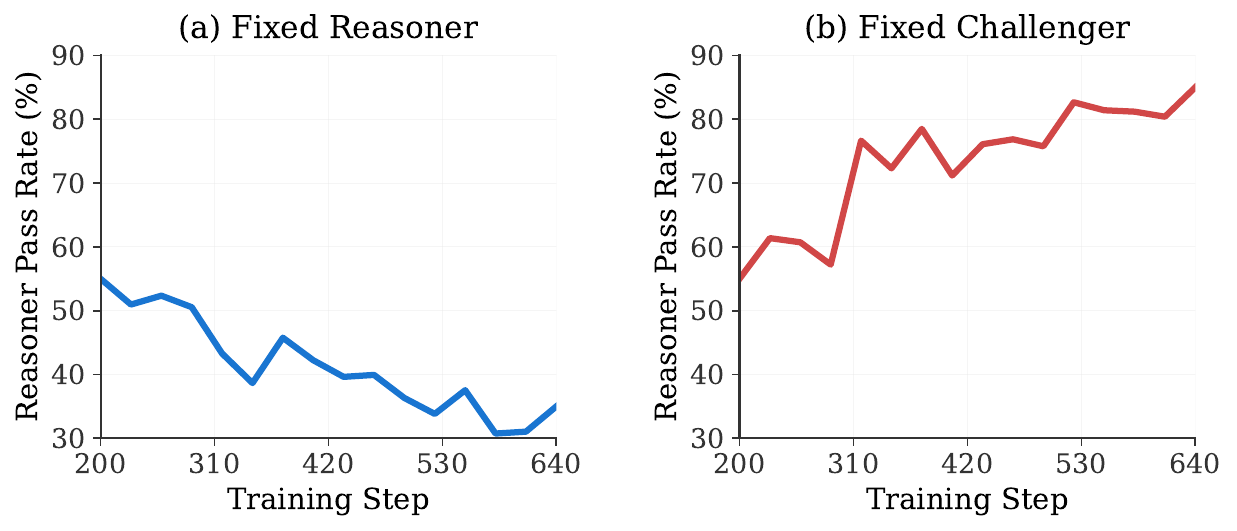}
    \caption{Reasoner pass rates when evaluating \modelname{} checkpoints at steps 200-640 against a fixed step-200 checkpoint on 128 documents. \textbf{(a) Fixed Reasoner:} Pass rate decreases from 55\% to 35\% as later \C{} checkpoints generate harder questions. \textbf{(b) Fixed Challenger:} Pass rate increases from 55\% to 85\% as later \R{} checkpoints improve at solving questions.}
    \label{fig:pass_rates}
    \end{minipage}
    \hfill
    \raisebox{0.1em}
    {\begin{minipage}[b]{0.48\textwidth}
    \centering
    \includegraphics[width=1.0\linewidth]{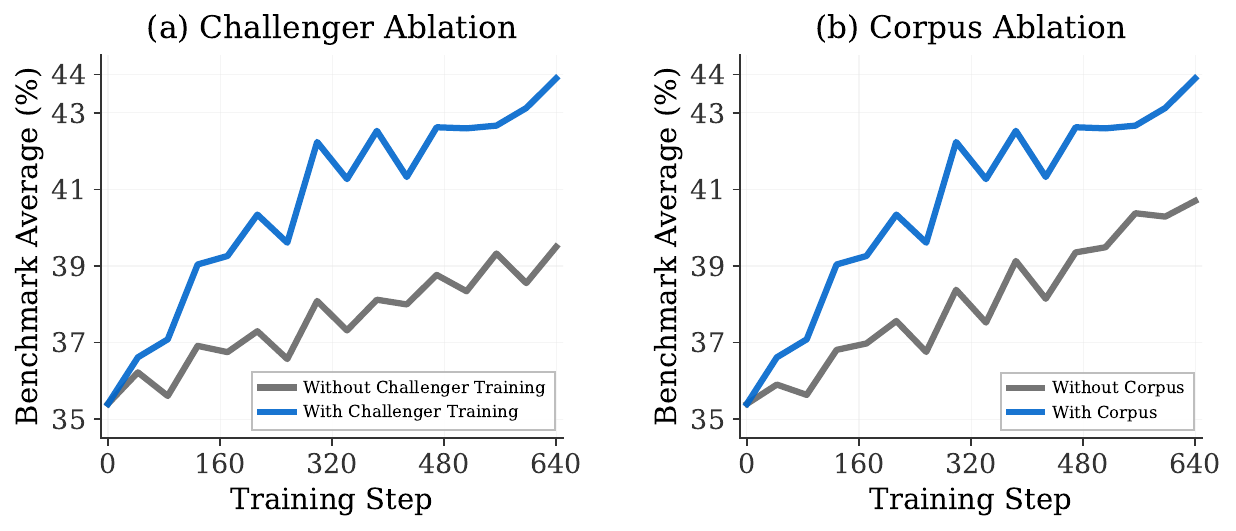}
    \caption{Training dynamics comparing key components of \modelname{}. \textbf{(a) Challenger Ablation:} Without \C{} training (fixed \C{}), performance improves less than with full adversarial training. \textbf{(b) Corpus Ablation:} With corpus grounding, the model achieves steady improvement to 43.9\%, while without it performance is lower at 40.7\%, showing the importance of external grounding.}
    \label{fig:training_dynamics}
    \end{minipage}
    }
\end{figure}

\textbf{Adversarial Learning Dynamics.} \autoref{fig:pass_rates} illustrates the co-evolution of \C{} and \R{} capabilities. To analyze the contribution of each role, we take checkpoints from full self-play training (steps 200-640) and evaluate them against a fixed step-200 checkpoint on a pool of 128 documents (with 128 generation attempts each). When evaluating different \C{} checkpoints against a fixed step-200 \R{} (subplot a), we observe the \R{}'s pass rate decreases from 55\% to 35\% as later \C{} checkpoints generate increasingly challenging questions. Conversely, when evaluating different \R{} checkpoints against a fixed step-200 \C{} (subplot b), the \R{}'s pass rate increases from 55\% to 85\% as later checkpoints become better at solving questions. In full \modelname{} training where both roles evolve simultaneously, this adversarial dynamic drives mutual improvement beyond what either role achieves in isolation.

\textbf{Challenger Learning Impact.} 
An important component of \modelname{} is learning the \C{} simultaneously with the \R{}, as opposed to using a fixed \C{}. Figure~\ref{fig:training_dynamics}(a) demonstrates that co-training the \C{} alongside the \R{} is essential for maximizing gains, validating our co-evolutionary approach. Without training the \C{}, the \R{} isn't challenged enough and improves more slowly.

\textbf{Corpus Grounding Impact.}  Finally, we examine the critical role of corpus grounding in \modelname{}. Figure~\ref{fig:training_dynamics}(b) illustrates the results. With corpus grounding, performance reaches 43.9\% through continuous access to diverse document contexts that provide near-inexhaustible question material. Without access to external documents, the model performance is lower at 40.7\%, indicating the importance of corpus grounding.

\subsection{Qualitative Analysis}

To better understand \modelname{}'s learning dynamics, we analyze the evolution of tasks and reasoning patterns produced by the \C{} and \R{}, respectively, throughout training.

\textbf{Challenger Task Development.} Figure~\ref{fig:task_examples} shows how task complexity evolves when the \C{} is given the same document at different training steps. Early tasks focus on surface-level information, while later tasks require deep comprehension and multi-step reasoning. At earlier steps, the \C{} cannot propose tasks that are too difficult, as the \R{} will not be able to learn from them. As training progresses, the \C{} learns to generate more difficult tasks to adapt to the improved \R{}'s capabilities.

\begin{figure}[h]
\centering
\begin{tcolorbox}[
    colframe=lightred,
    colback=verylightred,
    title=Early Training (Step 50): Surface-level Question,
    fonttitle=\bfseries\small
]
\scriptsize
\textbf{Document:} [Passage about solar system scale] "The diameter of the Sun is 1,391,000 km. The diameter of the Moon is 3,475 km. The distance from Earth to the Sun is 149,600,000 km. For a solar eclipse to occur, the Moon must appear the same angular size as the Sun from Earth. In a scale model, a circle with diameter 2 cm represents Earth. When Earth is scaled to 0.3 mm (like a period), the entire solar system model changes dramatically..."\\
\textbf{Generated MCQ:} What is the diameter of the Moon?\\
A) 1,391 km\\
B) 3,475 km\\
C) 34,750 km\\
D) 347,500 km\\
\textbf{Answer:} B
\end{tcolorbox}
\begin{tcolorbox}[
    colframe=mediumred,
    colback=verylightred,
    title=Late Training (Step 480): Multi-step Reasoning,
    fonttitle=\bfseries\small
]
\scriptsize
\textbf{Document:} [Passage about solar system scale] "The diameter of the Sun is 1,391,000 km. The diameter of the Moon is 3,475 km. The distance from Earth to the Sun is 149,600,000 km. For a solar eclipse to occur, the Moon must appear the same angular size as the Sun from Earth. In a scale model, a circle with diameter 2 cm represents Earth. When Earth is scaled to 0.3 mm (like a period), the entire solar system model changes dramatically..."\\
\textbf{Generated MCQ:} An alien civilization has a moon with diameter 3,475 km that creates perfect solar eclipses. Their star has the same diameter as our Sun. If their moon orbits at 374,000 km (maintaining the same angular size ratio as Earth's eclipse geometry), approximately what would be the distance from their planet to their star in kilometers?\\
A) 374,000,000 km\\
B) 149,600,000 km\\
C) 1,391,000,000 km\\
D) 74,800,000 km\\
\textbf{Answer:} B
\end{tcolorbox}
\caption{Evolution of the \C{}'s task complexity on the same document. The \C{} progresses from extracting explicit facts to generating questions requiring understanding of angular size relationships and proportional reasoning to arrive at document-stated values.}
\label{fig:task_examples}
\end{figure}

\textbf{Reasoner Pattern Development.} Figure~\ref{fig:reasoning_evolution} illustrates the evolution of the \R{}'s structured problem-solving approach given the same question at different training steps. As training progresses, increasingly sophisticated reasoning patterns emerge, reflecting the \R{}'s adaptation to the progressively more challenging tasks generated by the \C{}. This progression reveals authentic reasoning capabilities rather than mere memorization, as the model learns to systematically decompose problems, validate intermediate steps, and self-correct when inconsistencies arise.

\begin{figure}[ht]
\vspace{-5pt}
\centering
\begin{tcolorbox}[
    colframe=lightred,
    colback=verylightred,
    title=Early Training: Unstructured Reasoning,
    fonttitle=\bfseries\small
]
\scriptsize
\textbf{Question:} An alien civilization has a moon with diameter 3,475 km that creates perfect solar eclipses. Their star has the same diameter as our Sun. If their moon orbits at 374,000 km (maintaining the same angular size ratio as Earth's eclipse geometry), approximately what would be the distance from their planet to their star in kilometers?\\
A) 374,000,000 km  B) 149,600,000 km  C) 1,391,000,000 km  D) 74,800,000 km\\
\textbf{Reasoner Output:} ``Perfect eclipses mean moon and star look same size. Moon is at 374,000 km. Stars are way farther than moons, so maybe 1000 times farther? That gives 374,000,000 km. The answer is \boxed{A) 374,000,000 km}.''
\end{tcolorbox}
\begin{tcolorbox}[
    colframe=mediumred,
    colback=verylightred,
    title=Late Training: Structured Multi-step Reasoning,
    fonttitle=\bfseries\small
]
\scriptsize
\textbf{Question:} An alien civilization has a moon with diameter 3,475 km that creates perfect solar eclipses. Their star has the same diameter as our Sun. If their moon orbits at 374,000 km (maintaining the same angular size ratio as Earth's eclipse geometry), approximately what would be the distance from their planet to their star in kilometers?\\
A) 374,000,000 km  B) 149,600,000 km  C) 1,391,000,000 km  D) 74,800,000 km\\
\textbf{Reasoner Output:} ``Let me analyze this step-by-step.\\
\\
\#\# Step 1: Identify given information\\
- Alien moon diameter = 3,475 km\\
- Alien star diameter = same as Sun = 1,391,000 km\\
- Moon orbital distance = 374,000 km\\
- Need: star distance for perfect eclipses\\
\\
\#\# Step 2: Understand perfect eclipse condition\\
For perfect eclipses, angular sizes must be equal:\\
$\frac{\text{Moon diameter}}{\text{Moon distance}} = \frac{\text{Star diameter}}{\text{Star distance}}$\\
\\
\#\# Step 3: Set up the equation\\
$\frac{3,475}{374,000} = \frac{1,391,000}{d}$\\
\\
\#\# Step 4: Solve for star distance d\\
Cross multiply:\\
$3,475 \times d = 1,391,000 \times 374,000$\\
$d = \frac{1,391,000 \times 374,000}{3,475}$\\
\\
\#\# Step 5: Calculate\\
$d = \frac{520,234,000,000}{3,475} = 149,708,489$ km\\
\\
\#\# Step 6: Match to closest option\\
149,708,489 km $\approx$ 149,600,000 km\\
\\
\#\# Step 7: Verify\\
Moon angular size: $\frac{3,475}{374,000} = 0.00929$\\
Star angular size: $\frac{1,391,000}{149,600,000} = 0.00930$\\
These match (small rounding difference), confirming our solution.\\
\\
The answer is \boxed{B) 149,600,000 km}''
\end{tcolorbox}
\vspace{-5pt}
\caption{Evolution of \R{}'s reasoning patterns. The \R{} progresses from intuitive guessing to systematic application of angular size principles, proper ratio setup, algebraic manipulation, and solution verification.}
\label{fig:reasoning_evolution}
\end{figure}

\section{Ablations}
\label{sec:ablations}

To understand the key components driving \modelname{}'s performance, we conduct comprehensive ablation studies on Qwen3-4B-Base. We examine three critical design choices: corpus composition, task type distribution, and \C{} reward strategies. All experiments use the same training configuration as our main experiments.

\subsection{Corpus Distribution}
\label{sec:ablation_corpus_distribution}

One of the main ingredients of \modelname{} is the external corpora of documents. Compared to no corpora (pure ungrounded self-play), \modelname{} achieves substantial gains on both math and general reasoning tasks. However, we include two large document datasets: NaturalReasoning and Nemotron-CC-Math. In order to understand the effects of the document types, we ablate each document dataset separately. We show the results in ~\autoref{tab:ablation_corpus}. As expected, the NaturalReasoning dataset leads to the largest gains in general reasoning tasks as it contains documents targeted toward such tasks. Similarly, Nemotron-CC-Math leads to the biggest improvements in the math tasks. However, both datasets combined lead to the best overall performance.

\begin{table}[h!]
\centering
\small
\caption{Effect of corpus composition, used by the \C{}, on reasoning improvements. We show average performance across mathematical reasoning benchmarks (Math), general reasoning benchmarks (General), and their overall average. Best results shown in \textbf{bold}, second best in \underline{underline}. Full benchmark results are given in Appendix Table~\ref{tab:ablation_full}.}
\label{tab:ablation_corpus}
\begin{tabular}{lccc}
\toprule
\textbf{Corpus Type} & \textbf{Math Avg} & \textbf{General Avg} & \textbf{Overall Avg}  \\
\midrule
NaturalReasoning  & 44.4 & \textbf{37.0} & 41.7 \\
Nemotron-CC-Math  & \textbf{53.4} & 29.8 & 43.2 \\
\rowcolor{red!10}NaturalReasoning + Nemotron-CC-Math & \underline{50.6} & \underline{35.0} & \textbf{44.9} \\
\bottomrule
\end{tabular}
\end{table}

\subsection{Task Type}
\label{sec:ablation_task_type}
A second key feature of \modelname{} is the ability to generate different task types dependent on the current document. \modelname{} uses two different task types: multiple-choice (MCQ) and free-form. \autoref{tab:ablation_format} shows the results of only using MCQ or free-form versus using both. While using free-form questions only leads to the highest math gains, combining MCQ and free-form (mixing tasks) achieves the best overall performance. We attribute this to MCQs providing reliable verification while free-form questions encourage flexible reasoning.

\begin{table}[h]
\centering
\small
\caption{Effect of the task type, proposed by the \C{}, on reasoning improvements. MCQ provides structured answer choices while free-form requires open-ended generation. Combining both leads to the best overall performance. Full results are provided in Appendix  Table~\ref{tab:ablation_format_full}.}
\label{tab:ablation_format}
\begin{tabular}{lccc}
\toprule
\textbf{Task Type} & \textbf{Math Avg} & \textbf{General Avg} & \textbf{Overall Avg}  \\
\midrule
MCQ only & 46.9 & \textbf{35.7} & 42.0 \\
Free-form only & \textbf{52.5} & 31.8 & 43.7 \\
\rowcolor{red!10}MCQ + Free-form & \underline{50.6} & \underline{35.0} & \textbf{44.9} \\
\bottomrule
\end{tabular}
\end{table}

\subsection{Challenger's Reward}
\label{sec:ablation_challenger_reward}

Lastly, we find that the \C{}'s reward function critically influences task difficulty calibration. We compare four reward strategies that shape how a \C{} can select problems during self-play with a \R{} (see Appendix~\ref{app:reward_functions}, \autoref{fig:reward_functions} for reward visualizations).
Absolute Zero rewards the \C{} with one minus the \R{}'s average success rate: tasks where the \R{} fails more often receive higher reward. While intuitive, this task conflates difficulty with learning value. A threshold reward over number of correct rollouts uses a binary signal: reward 1 for relatively solvable tasks, 0 for tasks with 0\% or 100\% pass rate (impossible or trivial). The R-Zero reward explicitly targets maximum uncertainty: it rewards tasks where the \R{}'s multiple responses split evenly between different answers, peaking when exactly half agree with the most common answer. Our variance approach uses a related principle, but captures the full spread of the answer distribution across multiple \R{} samples rather than measuring agreement with a single mode, creating a curriculum calibrated at the frontier of the model's capability.
Results, given in \autoref{tab:ablation_challenger_reward}, show 
\modelname{}'s variance-based reward achieves the best performance across all metrics. A critical insight is that optimal learning occurs when the \R{} has a balanced success rate.

\begin{table}[h]
\centering
\small
\caption{Effect of the \C{} reward strategies on reasoning performance. Each method targets different aspects of task difficulty for curriculum generation. Our Variance reward outperforms previously proposed rewards.}
\label{tab:ablation_challenger_reward}
\begin{tabular}{lccc}
\toprule
\textbf{Challenger Reward} & \textbf{Math} & \textbf{General} & \textbf{Overall}  \\
\midrule
Absolute Zero  & 48.2 & 30.8 & 40.7  \\
Threshold   & 48.6 & 31.6 & 41.4  \\
R-Zero & \underline{50.0} & \underline{33.9} & \underline{43.6}  \\
\rowcolor{red!10}Variance (SPICE) & \textbf{50.6} & \textbf{35.0} & \textbf{44.9} \\
\bottomrule
\end{tabular}
\end{table}

\section{Related Work}

\textbf{Reinforcement Learning for LLM Reasoning\,}
Reinforcement learning has evolved from alignment tasks using RLHF~\citep{jaques2019way,ouyang2022training,bai2022constitutional} to directly improving reasoning capabilities. Recent models like OpenAI o1~\citep{openai2024o1} and DeepSeek-R1~\citep{deepseek2024r1} demonstrate that RL with verifiable rewards (RLVR) can unlock chain-of-thought reasoning using rule-based rewards~\citep{lightman2023let,uesato2022solving}. However, these approaches depend on human-curated problem sets and domain-specific reward engineering, limiting their scalability. The Era of Experience~\citep{silver2025era} argues that future AI systems must learn from environmental interaction rather than static datasets~\citep{jin2025era}. Yet, current RLVR methods remain bound to pre-collected problems. \modelname{} aims to address this by generating problems dynamically from corpus interactions, enabling learning without human curation.

\textbf{Multi-Agent RL for Language Models\,}
Implementing multi-agent RL for full-scale LLMs presents significant technical challenges~\citep{wan2025rema,liu2025chasingmovingtargetsonline}. Prior work has made various compromises: \citet{sarkar2025training} uses RNNs instead of transformers; \citet{jacob2022emergent} focuses on simplified environments that don't require full autoregressive generation; and \citet{liao2024efficacylanguagemodelselfplay} shows the effectiveness of self-play with SFT on proprietary models. SPIRAL~\citep{liu2025spiral} demonstrates that zero-sum games between agents can teach transferable reasoning through multi-turn interactions, but requires carefully designed game environments. \modelname{} builds on multi-agent frameworks but grounds the adversarial dynamics in corpus contexts, enabling automatic curriculum emergence without engineered environments.

\textbf{Self-Play for Autonomous Improvement\,}
Self-play has revolutionized game-playing AI, including TD-Gammon's backgammon mastery~\citep{tesauro1995temporal}, AlphaGo's superhuman Go performance~\citep{silver2017mastering}, and CICERO's comprehension of cooperative strategies \citep{meta2022human}. Self-play can even create automatic curricula through intrinsic motivation~\citep{sukhbaatar2018intrinsicmotivationautomaticcurricula}. In language models, self-play began with alignment methods like SPIN~\citep{chen2024self} and Self-Rewarding Language Models~\citep{yuan2401self}. Recent work explores capability improvement: SPAG~\citep{cheng2024self} applies self-play to Adversarial Taboo but uses offline updates and remains confined to a single word game; SPC~\citep{chen2025spcevolvingselfplaycritic} and Genius~\citep{xu2025geniusgeneralizablepurelyunsupervised} require human task distributions as seeds; SPELL~\citep{yang2025spell} targets long-context evolution through self-play; Language Self-Play~\citep{kuba2025language} explores data-free training paradigms. Pure self-play methods consistently demonstrate empirical limitations: R-Zero~\citep{huang2025r} achieves initial gains but degrades after 3-4 iterations with pseudo-label accuracy dropping from 79\% to 63\%; Absolute Zero~\citep{zhao2025absolute} and SQLM~\citep{chen2025self} generate coding tasks and hence remain limited in domain.

\textbf{Synthetic Question Generation from Corpora\,}
Synthetic question generation methods either bootstrap from existing datasets or mine from corpora. Bootstrapping approaches like STaR~\citep{zelikman2022star}, MetaMath~\citep{yu2023metamath}, and SvS~\citep{liang2025beyond} remain bounded by initial dataset coverage. Self-Instruct \citep{wang2022self} and reasoning-based CoT-self-instruct~\citep{yu2025cot} generate high-quality synthetic prompts from few-shot examples, but create static datasets rather than adaptive curricula. Corpus-mining methods achieve broader coverage: WebInstruct~\citep{yue2024mammoth2} harvests QA pairs but requires rule-based filters; General-Reasoner~\citep{ma2025general} and NaturalReasoning~\citep{yuan2025naturalreasoning} generate millions of questions from web content but create static offline datasets. \modelname{} differs fundamentally through online adversarial generation where the \C{} continuously mines corpus contexts to generate tasks calibrated to the Reasoner's current capability, preventing both coverage limitations and quality degradation.

\section{Conclusion}

We introduced \modelname{}, a self-play RL framework that aims to overcome the issues of hallucination and information symmetry, outperforming pure (ungrounded) self-play. By treating a large document corpus as a near-inexhaustible external environment, \modelname{} presents a new paradigm for self-improvement, allowing LLMs to interact with a continually changing world via web documents. Through adversarial dynamics between a \C{} that generates tasks and a \R{} that solves them, our system creates its own ever more challenging goals and strives to achieve them. \modelname{} achieves strong improvements across mathematical and general reasoning benchmarks compared to state-of-the-art self-play methods. By mining training signals from the vast knowledge embedded in corpora, \modelname{} develops reasoning capabilities that transfer broadly across domains. We believe this shift from closed-loop self-play to corpus-grounded adversarial learning opens new avenues for self-improvement without explicit human supervision.

\clearpage
\newpage
\bibliographystyle{assets/plainnat}
\bibliography{paper}

\begin{thebibliography}{65}
\providecommand{\natexlab}[1]{#1}
\providecommand{\url}[1]{\texttt{#1}}
\expandafter\ifx\csname urlstyle\endcsname\relax
  \providecommand{\doi}[1]{doi: #1}\else
  \providecommand{\doi}{doi: \begingroup \urlstyle{rm}\Url}\fi

\bibitem[Achiam et~al.(2023)Achiam, Adler, Agarwal, Ahmad, Akkaya, Aleman, Almeida, Altenschmidt, Altman, Anadkat, et~al.]{achiam2023gpt}
Josh Achiam, Steven Adler, Sandhini Agarwal, Lama Ahmad, Ilge Akkaya, Florencia~Leoni Aleman, Diogo Almeida, Janko Altenschmidt, Sam Altman, Shyamal Anadkat, et~al.
\newblock Gpt-4 technical report.
\newblock \emph{arXiv preprint arXiv:2303.08774}, 2023.

\bibitem[Bai et~al.(2022)Bai, Kadavath, Kundu, Askell, Kernion, Jones, Chen, Goldie, Mirhoseini, McKinnon, et~al.]{bai2022constitutional}
Yuntao Bai, Saurav Kadavath, Sandipan Kundu, Amanda Askell, Jackson Kernion, Andy Jones, Anna Chen, Anna Goldie, Azalia Mirhoseini, Cameron McKinnon, et~al.
\newblock Constitutional ai: Harmlessness from ai feedback.
\newblock \emph{arXiv preprint arXiv:2212.08073}, 2022.

\bibitem[Chen et~al.(2025{\natexlab{a}})Chen, Zhang, Ma, Wang, Liang, Tu, Li, and Wong]{chen2025spcevolvingselfplaycritic}
Jiaqi Chen, Bang Zhang, Ruotian Ma, Peisong Wang, Xiaodan Liang, Zhaopeng Tu, Xiaolong Li, and Kwan-Yee~K Wong.
\newblock Spc: Evolving self-play critic via adversarial games for llm reasoning.
\newblock \emph{arXiv preprint arXiv:2504.19162}, 2025{\natexlab{a}}.

\bibitem[Chen et~al.(2025{\natexlab{b}})Chen, Prabhudesai, Fragkiadaki, Liu, and Pathak]{chen2025self}
Lili Chen, Mihir Prabhudesai, Katerina Fragkiadaki, Hao Liu, and Deepak Pathak.
\newblock Self-questioning language models.
\newblock \emph{arXiv preprint arXiv:2508.03682}, 2025{\natexlab{b}}.

\bibitem[Chen et~al.(2024)Chen, Deng, Yuan, Ji, and Gu]{chen2024self}
Zixiang Chen, Yihe Deng, Huizhuo Yuan, Kaixuan Ji, and Quanquan Gu.
\newblock Self-play fine-tuning converts weak language models to strong language models.
\newblock In \emph{ICML}, 2024.

\bibitem[Cheng et~al.(2024)Cheng, Hu, Xu, Zhang, Dai, Han, Li, et~al.]{cheng2024self}
Pengyu Cheng, Tianhao Hu, Han Xu, Zhisong Zhang, Yong Dai, Lei Han, Xiaolong Li, et~al.
\newblock Self-playing adversarial language game enhances llm reasoning.
\newblock \emph{Advances in Neural Information Processing Systems}, 37:\penalty0 126515--126543, 2024.

\bibitem[Clune(2019)]{clune2019ai}
Jeff Clune.
\newblock Ai-gas: Ai-generating algorithms, an alternate paradigm for producing general artificial intelligence.
\newblock \emph{arXiv preprint arXiv:1905.10985}, 2019.

\bibitem[Cobbe et~al.(2021)Cobbe, Kosaraju, Bavarian, Chen, Jun, Kaiser, Plappert, Tworek, Hilton, Nakano, Hesse, and Schulman]{cobbe2021gsm8k}
Karl Cobbe, Vineet Kosaraju, Mohammad Bavarian, Mark Chen, Heewoo Jun, Lukasz Kaiser, Matthias Plappert, Jerry Tworek, Jacob Hilton, Reiichiro Nakano, Christopher Hesse, and John Schulman.
\newblock Training verifiers to solve math word problems, 2021.
\newblock \url{https://arxiv.org/abs/2110.14168}.

\bibitem[{DeepSeek Team}(2024)]{deepseek2024r1}
{DeepSeek Team}.
\newblock Deepseek-r1: Incentivizing reasoning capability in llms via reinforcement learning.
\newblock \emph{arXiv preprint arXiv:2401.00000}, 2024.

\bibitem[Du et~al.(2025)Du, Yao, Ma, Wang, Zheng, Zhu, Liu, Liang, Jin, Wei, et~al.]{du2025supergpqa}
Xinrun Du, Yifan Yao, Kaijing Ma, Bingli Wang, Tianyu Zheng, King Zhu, Minghao Liu, Yiming Liang, Xiaolong Jin, Zhenlin Wei, et~al.
\newblock Supergpqa: Scaling llm evaluation across 285 graduate disciplines.
\newblock \emph{arXiv preprint arXiv:2502.14739}, 2025.

\bibitem[Dubey et~al.(2024)Dubey, Jauhri, Pandey, Kadian, Al-Dahle, Letman, Mathur, Schelten, Yang, Fan, et~al.]{dubey2024llama}
Abhimanyu Dubey, Abhinav Jauhri, Abhinav Pandey, Abhishek Kadian, Ahmad Al-Dahle, Aiesha Letman, Akhil Mathur, Alan Schelten, Amy Yang, Angela Fan, et~al.
\newblock The llama 3 herd of models.
\newblock \emph{arXiv e-prints}, pages arXiv--2407, 2024.

\bibitem[Espeholt et~al.(2018)Espeholt, Soyer, Munos, Simonyan, Mnih, Ward, Doron, Firoiu, Harley, Dunning, et~al.]{espeholt2018impala}
Lasse Espeholt, Hubert Soyer, Remi Munos, Karen Simonyan, Vlad Mnih, Tom Ward, Yotam Doron, Vlad Firoiu, Tim Harley, Iain Dunning, et~al.
\newblock Impala: Scalable distributed deep-rl with importance weighted actor-learner architectures.
\newblock In \emph{International conference on machine learning}, pages 1407--1416. PMLR, 2018.

\bibitem[FAIR et~al.(2022)FAIR, Bakhtin, Brown, Dinan, Farina, Flaherty, Fried, Goff, Gray, Hu, et~al.]{meta2022human}
FAIR, Anton Bakhtin, Noam Brown, Emily Dinan, Gabriele Farina, Colin Flaherty, Daniel Fried, Andrew Goff, Jonathan Gray, Hengyuan Hu, et~al.
\newblock Human-level play in the game of diplomacy by combining language models with strategic reasoning.
\newblock \emph{Science}, 378\penalty0 (6624):\penalty0 1067--1074, 2022.

\bibitem[He et~al.(2024)He, Luo, Bai, Hu, Thai, Shen, Hu, Han, Huang, Zhang, Liu, Qi, Liu, and Sun]{he2024olympiadbench}
Chaoqun He, Renjie Luo, Yuzhuo Bai, Shengding Hu, Zhen~Leng Thai, Junhao Shen, Jinyi Hu, Xu~Han, Yujie Huang, Yuxiang Zhang, Jie Liu, Lei Qi, Zhiyuan Liu, and Maosong Sun.
\newblock Olympiadbench: A challenging benchmark for promoting agi with olympiad-level bilingual multimodal scientific problems, 2024.

\bibitem[Hendrycks et~al.(2021)Hendrycks, Burns, Kadavath, Arora, Basart, Tang, Song, and Steinhardt]{hendrycksmath2021}
Dan Hendrycks, Collin Burns, Saurav Kadavath, Akul Arora, Steven Basart, Eric Tang, Dawn Song, and Jacob Steinhardt.
\newblock Measuring mathematical problem solving with the math dataset.
\newblock \emph{NeurIPS}, 2021.

\bibitem[Huang et~al.(2025)Huang, Yu, Wang, Zhang, Li, Li, Huang, Mi, and Yu]{huang2025r}
Chengsong Huang, Wenhao Yu, Xiaoyang Wang, Hongming Zhang, Zongxia Li, Ruosen Li, Jiaxin Huang, Haitao Mi, and Dong Yu.
\newblock R-zero: Self-evolving reasoning llm from zero data.
\newblock \emph{arXiv preprint arXiv:2508.05004}, 2025.

\bibitem[Jacob et~al.(2022)Jacob, Gupta, and Andreas]{jacob2022emergent}
Athul~Paul Jacob, Abhishek Gupta, and Jacob Andreas.
\newblock Emergent linguistic phenomena in multi-agent communication games.
\newblock \emph{arXiv preprint arXiv:2205.05984}, 2022.

\bibitem[Jaques et~al.(2019)Jaques, Ghandeharioun, Shen, Ferguson, Lapedriza, Jones, Gu, and Picard]{jaques2019way}
Natasha Jaques, Asma Ghandeharioun, Judy~Hanwen Shen, Craig Ferguson, Agata Lapedriza, Noah Jones, Shixiang Gu, and Rosalind Picard.
\newblock Way off-policy batch deep reinforcement learning of implicit human preferences in dialog.
\newblock \emph{arXiv preprint arXiv:1907.00456}, 2019.

\bibitem[Jin et~al.(2025)Jin, Xu, Liu, Tao, Golovneva, Shu, Zhao, Li, and Weston]{jin2025era}
Chuanyang Jin, Jing Xu, Bo~Liu, Leitian Tao, Olga Golovneva, Tianmin Shu, Wenting Zhao, Xian Li, and Jason Weston.
\newblock The era of real-world human interaction: Rl from user conversations.
\newblock \emph{arXiv preprint arXiv:2509.25137}, 2025.

\bibitem[Kaplan et~al.(2020)Kaplan, McCandlish, Henighan, Brown, Chess, Child, Gray, Radford, Wu, and Amodei]{kaplan2020scaling}
Jared Kaplan, Sam McCandlish, Tom Henighan, Tom~B Brown, Benjamin Chess, Rewon Child, Scott Gray, Alec Radford, Jeffrey Wu, and Dario Amodei.
\newblock Scaling laws for neural language models.
\newblock \emph{arXiv preprint arXiv:2001.08361}, 2020.

\bibitem[Kazemi et~al.(2025)Kazemi, Fatemi, Bansal, Palowitch, Anastasiou, Mehta, Jain, Aglietti, Jindal, Chen, Dikkala, Tyen, Liu, Shalit, Chiappa, Olszewska, Tay, Tran, Le, and Firat]{bbeh}
Mehran Kazemi, Bahare Fatemi, Hritik Bansal, John Palowitch, Chrysovalantis Anastasiou, Sanket~Vaibhav Mehta, Lalit~K. Jain, Virginia Aglietti, Disha Jindal, Peter Chen, Nishanth Dikkala, Gladys Tyen, Xin Liu, Uri Shalit, Silvia Chiappa, Kate Olszewska, Yi~Tay, Vinh~Q. Tran, Quoc~V. Le, and Orhan Firat.
\newblock Big-bench extra hard, 2025.
\newblock \url{https://arxiv.org/abs/2502.19187}.

\bibitem[Kojima et~al.(2022)Kojima, Gu, Reid, Matsuo, and Iwasawa]{kojima2022large}
Takeshi Kojima, Shixiang~Shane Gu, Machel Reid, Yutaka Matsuo, and Yusuke Iwasawa.
\newblock Large language models are zero-shot reasoners.
\newblock \emph{Advances in neural information processing systems}, 35:\penalty0 22199--22213, 2022.

\bibitem[Kuba et~al.(2025)Kuba, Gu, Ma, Tian, and Mohan]{kuba2025language}
Jakub~Grudzien Kuba, Mengting Gu, Qi~Ma, Yuandong Tian, and Vijai Mohan.
\newblock Language self-play for data-free training.
\newblock \emph{arXiv preprint arXiv:2509.07414}, 2025.

\bibitem[Kwon et~al.(2023)Kwon, Li, Zhuang, Sheng, Zheng, Yu, Gonzalez, Zhang, and Stoica]{kwon2023efficient}
Woosuk Kwon, Zhuohan Li, Siyuan Zhuang, Ying Sheng, Lianmin Zheng, Cody~Hao Yu, Joseph~E. Gonzalez, Hao Zhang, and Ion Stoica.
\newblock Efficient memory management for large language model serving with pagedattention.
\newblock In \emph{Proceedings of the ACM SIGOPS 29th Symposium on Operating Systems Principles}, 2023.

\bibitem[Lewkowycz et~al.(2022)Lewkowycz, Andreassen, Dohan, Dyer, Michalewski, Ramasesh, Slone, Anil, Schlag, Gutman-Solo, et~al.]{lewkowycz2022solving}
Aitor Lewkowycz, Anders Andreassen, David Dohan, Ethan Dyer, Henryk Michalewski, Vinay Ramasesh, Ambrose Slone, Cem Anil, Imanol Schlag, Theo Gutman-Solo, et~al.
\newblock Solving quantitative reasoning problems with language models.
\newblock \emph{Advances in Neural Information Processing Systems}, 35:\penalty0 3843--3857, 2022.

\bibitem[Li et~al.(2024)Li, Fang, Smyrnis, Ivgi, Jordan, Gadre, Bansal, Guha, Keh, Arora, et~al.]{li2024datacomp}
Jeffrey Li, Alex Fang, Georgios Smyrnis, Maor Ivgi, Matt Jordan, Samir~Yitzhak Gadre, Hritik Bansal, Etash Guha, Sedrick~Scott Keh, Kushal Arora, et~al.
\newblock Datacomp-lm: In search of the next generation of training sets for language models.
\newblock \emph{Advances in Neural Information Processing Systems}, 37:\penalty0 14200--14282, 2024.

\bibitem[Liang et~al.(2025)Liang, Li, Gong, Shen, Wu, Guo, and Chen]{liang2025beyond}
Xiao Liang, Zhongzhi Li, Yeyun Gong, Yelong Shen, Ying~Nian Wu, Zhijiang Guo, and Weizhu Chen.
\newblock Beyond pass@ 1: Self-play with variational problem synthesis sustains rlvr.
\newblock \emph{arXiv preprint arXiv:2508.14029}, 2025.

\bibitem[Liao et~al.(2024)Liao, Tomlin, and Klein]{liao2024efficacylanguagemodelselfplay}
Austen Liao, Nicholas Tomlin, and Dan Klein.
\newblock Efficacy of language model self-play in non-zero-sum games, 2024.
\newblock \url{https://arxiv.org/abs/2406.18872}.

\bibitem[Lightman et~al.(2023)Lightman, Kosaraju, Burda, Edwards, Baker, Lee, Leike, Schulman, Sutskever, and Cobbe]{lightman2023let}
Hunter Lightman, Vineet Kosaraju, Yuri Burda, Harri Edwards, Bowen Baker, Teddy Lee, Jan Leike, John Schulman, Ilya Sutskever, and Karl Cobbe.
\newblock Let's verify step by step.
\newblock \emph{arXiv preprint arXiv:2305.20050}, 2023.

\bibitem[Liu et~al.(2025{\natexlab{a}})Liu, Guertler, Yu, Liu, Qi, Balcells, Liu, Tan, Shi, Lin, et~al.]{liu2025spiral}
Bo~Liu, Leon Guertler, Simon Yu, Zichen Liu, Penghui Qi, Daniel Balcells, Mickel Liu, Cheston Tan, Weiyan Shi, Min Lin, et~al.
\newblock Spiral: Self-play on zero-sum games incentivizes reasoning via multi-agent multi-turn reinforcement learning.
\newblock \emph{arXiv preprint arXiv:2506.24119}, 2025{\natexlab{a}}.

\bibitem[Liu et~al.(2025{\natexlab{b}})Liu, Jiang, Liang, Du, Choi, Althoff, and Jaques]{liu2025chasingmovingtargetsonline}
Mickel Liu, Liwei Jiang, Yancheng Liang, Simon~Shaolei Du, Yejin Choi, Tim Althoff, and Natasha Jaques.
\newblock Chasing moving targets with online self-play reinforcement learning for safer language models, 2025{\natexlab{b}}.
\newblock \url{https://arxiv.org/abs/2506.07468}.

\bibitem[Liu et~al.(2024)Liu, Chen, Wan, Du, Lee, and Lin]{liu2024oat}
Zichen Liu, Changyu Chen, Xinyi Wan, Chao Du, Wee~Sun Lee, and Min Lin.
\newblock Oat: A research-friendly framework for llm online alignment.
\newblock \url{https://github.com/sail-sg/oat}, 2024.

\bibitem[Liu et~al.(2025{\natexlab{c}})Liu, Chen, Li, Qi, Pang, Du, Lee, and Lin]{liu2025understanding}
Zichen Liu, Changyu Chen, Wenjun Li, Penghui Qi, Tianyu Pang, Chao Du, Wee~Sun Lee, and Min Lin.
\newblock Understanding r1-zero-like training: A critical perspective.
\newblock \emph{arXiv preprint arXiv:2503.20783}, 2025{\natexlab{c}}.

\bibitem[Ma et~al.(2025)Ma, Liu, Jiang, Zhang, Ma, and Chen]{ma2025general}
Xueguang Ma, Qian Liu, Dongfu Jiang, Ge~Zhang, Zejun Ma, and Wenhu Chen.
\newblock General-reasoner: Advancing llm reasoning across all domains.
\newblock \emph{arXiv preprint arXiv:2505.14652}, 2025.

\bibitem[MAA({\natexlab{a}})]{aime}
MAA.
\newblock American invitational mathematics examination ({AIME}).
\newblock Mathematics Competition Series, n.d.{\natexlab{a}}.
\newblock \url{https://maa.org/math-competitions/aime}.

\bibitem[MAA({\natexlab{b}})]{amc}
MAA.
\newblock American mathematics competitions ({AMC} 10/12).
\newblock Mathematics Competition Series, n.d.{\natexlab{b}}.
\newblock \url{https://maa.org/math-competitions/amc}.

\bibitem[Mahabadi et~al.(2025)Mahabadi, Satheesh, Prabhumoye, Patwary, Shoeybi, and Catanzaro]{mahabadi2025nemotron}
Rabeeh~Karimi Mahabadi, Sanjeev Satheesh, Shrimai Prabhumoye, Mostofa Patwary, Mohammad Shoeybi, and Bryan Catanzaro.
\newblock Nemotron-cc-math: A 133 billion-token-scale high quality math pretraining dataset.
\newblock \emph{arXiv preprint arXiv:2508.15096}, 2025.

\bibitem[Morris et~al.(2025)Morris, Sitawarin, Guo, Kokhlikyan, Suh, Rush, Chaudhuri, and Mahloujifar]{morris2025much}
John~X Morris, Chawin Sitawarin, Chuan Guo, Narine Kokhlikyan, G~Edward Suh, Alexander~M Rush, Kamalika Chaudhuri, and Saeed Mahloujifar.
\newblock How much do language models memorize?
\newblock \emph{arXiv preprint arXiv:2505.24832}, 2025.

\bibitem[OpenAI(2024)]{openai2024gpt4ocard}
OpenAI.
\newblock {GPT-4o} system card, 2024.
\newblock \url{https://arxiv.org/abs/2410.21276}.

\bibitem[{OpenAI}(2024)]{openai2024o1}
{OpenAI}.
\newblock Learning to reason with llms.
\newblock \emph{OpenAI Blog}, 2024.
\newblock \url{https://openai.com/o1}.

\bibitem[Ouyang et~al.(2022)Ouyang, Wu, Jiang, Almeida, Wainwright, Mishkin, Zhang, Agarwal, Slama, Ray, et~al.]{ouyang2022training}
Long Ouyang, Jeffrey Wu, Xu~Jiang, Diogo Almeida, Carroll Wainwright, Pamela Mishkin, Chong Zhang, Sandhini Agarwal, Katarina Slama, Alex Ray, et~al.
\newblock Training language models to follow instructions with human feedback.
\newblock \emph{Advances in Neural Information Processing Systems}, 35:\penalty0 27730--27744, 2022.

\bibitem[Rein et~al.(2023)Rein, Hou, Stickland, Petty, Pang, Dirani, Michael, and Bowman]{rein2023gpqagraduatelevelgoogleproofqa}
David Rein, Betty~Li Hou, Asa~Cooper Stickland, Jackson Petty, Richard~Yuanzhe Pang, Julien Dirani, Julian Michael, and Samuel~R. Bowman.
\newblock {GPQA}: A graduate-level google-proof q\&a benchmark, 2023.
\newblock \url{https://arxiv.org/abs/2311.12022}.

\bibitem[Sarkar et~al.(2025)Sarkar, Xia, Liu, and Sadigh]{sarkar2025training}
Bidipta Sarkar, Warren Xia, C~Karen Liu, and Dorsa Sadigh.
\newblock Training language models for social deduction with multi-agent reinforcement learning.
\newblock \emph{arXiv preprint arXiv:2502.06060}, 2025.

\bibitem[Schmidhuber(2007)]{schmidhuber2007godel}
J{\"u}rgen Schmidhuber.
\newblock G{\"o}del machines: Fully self-referential optimal universal self-improvers.
\newblock In \emph{Artificial general intelligence}, pages 199--226. Springer, 2007.

\bibitem[Silver and Sutton(2025)]{silver2025era}
David Silver and Richard~S. Sutton.
\newblock Welcome to the era of experience.
\newblock In \emph{Designing an Intelligence}. MIT Press, 2025.
\newblock Preprint.

\bibitem[Silver et~al.(2017)Silver, Hubert, Schrittwieser, Antonoglou, Lai, Guez, Lanctot, Sifre, Kumaran, Graepel, et~al.]{silver2017mastering}
David Silver, Thomas Hubert, Julian Schrittwieser, Ioannis Antonoglou, Matthew Lai, Arthur Guez, Marc Lanctot, Laurent Sifre, Dharshan Kumaran, Thore Graepel, et~al.
\newblock Mastering chess and shogi by self-play with a general reinforcement learning algorithm.
\newblock \emph{arXiv preprint arXiv:1712.01815}, 2017.

\bibitem[Sukhbaatar et~al.(2017)Sukhbaatar, Lin, Kostrikov, Synnaeve, Szlam, and Fergus]{sukhbaatar2018intrinsicmotivationautomaticcurricula}
Sainbayar Sukhbaatar, Zeming Lin, Ilya Kostrikov, Gabriel Synnaeve, Arthur Szlam, and Rob Fergus.
\newblock Intrinsic motivation and automatic curricula via asymmetric self-play, 2017.
\newblock \url{https://arxiv.org/abs/1703.05407}.

\bibitem[Tesauro(1995)]{tesauro1995temporal}
Gerald Tesauro.
\newblock Temporal difference learning and td-gammon.
\newblock \emph{Communications of the ACM}, 38\penalty0 (3):\penalty0 58--68, 1995.

\bibitem[Uesato et~al.(2022)Uesato, Kushman, Kumar, Song, Siegel, Wang, Creswell, Irving, and Higgins]{uesato2022solving}
Jonathan Uesato, Nate Kushman, Ramana Kumar, Francis Song, Noah Siegel, Lisa Wang, Antonia Creswell, Geoffrey Irving, and Irina Higgins.
\newblock Solving math word problems with process-and outcome-based feedback.
\newblock \emph{arXiv preprint arXiv:2211.14275}, 2022.

\bibitem[Wan et~al.(2025)Wan, Li, Wen, Song, Wang, Yang, Schmidt, Wang, Zhang, Hu, et~al.]{wan2025rema}
Ziyu Wan, Yunxiang Li, Xiaoyu Wen, Yan Song, Hanjing Wang, Linyi Yang, Mark Schmidt, Jun Wang, Weinan Zhang, Shuyue Hu, et~al.
\newblock Rema: Learning to meta-think for llms with multi-agent reinforcement learning.
\newblock \emph{arXiv preprint arXiv:2503.09501}, 2025.

\bibitem[Wang et~al.(2022)Wang, Kordi, Mishra, Liu, Smith, Khashabi, and Hajishirzi]{wang2022self}
Yizhong Wang, Yeganeh Kordi, Swaroop Mishra, Alisa Liu, Noah~A Smith, Daniel Khashabi, and Hannaneh Hajishirzi.
\newblock Self-instruct: Aligning language models with self-generated instructions.
\newblock \emph{arXiv preprint arXiv:2212.10560}, 2022.

\bibitem[Wang et~al.(2024)Wang, Ma, Zhang, Ni, Chandra, Guo, Ren, Arulraj, He, Jiang, et~al.]{wang2024mmlupro}
Yubo Wang, Xueguang Ma, Ge~Zhang, Yuansheng Ni, Abhranil Chandra, Shiguang Guo, Weiming Ren, Aaran Arulraj, Xuan He, Ziyan Jiang, et~al.
\newblock Mmlu-pro: A more robust and challenging multi-task language understanding benchmark.
\newblock \emph{Advances in Neural Information Processing Systems}, 37:\penalty0 95266--95290, 2024.

\bibitem[Wang et~al.(2025)Wang, Zhou, Li, and Liu]{wang2025octothinker}
Zengzhi Wang, Fan Zhou, Xuefeng Li, and Pengfei Liu.
\newblock Octothinker: Mid-training incentivizes reinforcement learning scaling.
\newblock \emph{arXiv preprint arXiv:2506.20512}, 2025.

\bibitem[Wei et~al.(2022)Wei, Wang, Schuurmans, Bosma, Ichter, Xia, Chi, Le, and Zhou]{wei2022chain}
Jason Wei, Xuezhi Wang, Dale Schuurmans, Maarten Bosma, Brian Ichter, Fei Xia, Ed~Chi, Quoc Le, and Denny Zhou.
\newblock Chain-of-thought prompting elicits reasoning in large language models.
\newblock In \emph{Advances in Neural Information Processing Systems}, volume~35, pages 24824--24837, 2022.

\bibitem[Xu et~al.(2025)]{xu2025geniusgeneralizablepurelyunsupervised}
Fangzhi Xu et~al.
\newblock Genius: A generalizable and purely unsupervised self-training framework for advanced reasoning, 2025.

\bibitem[Yang et~al.(2025{\natexlab{a}})Yang, Li, Yang, Zhang, Hui, Zheng, Yu, Gao, Huang, Lv, et~al.]{yang2025qwen3}
An~Yang, Anfeng Li, Baosong Yang, Beichen Zhang, Binyuan Hui, Bo~Zheng, Bowen Yu, Chang Gao, Chengen Huang, Chenxu Lv, et~al.
\newblock Qwen3 technical report.
\newblock \emph{arXiv preprint arXiv:2505.09388}, 2025{\natexlab{a}}.

\bibitem[Yang et~al.(2025{\natexlab{b}})Yang, Shen, Chen, Li, Wan, Yan, Quan, and Huang]{yang2025spell}
Ziyi Yang, Weizhou Shen, Ruijun Chen, Chenliang Li, Fanqi Wan, Ming Yan, Xiaojun Quan, and Fei Huang.
\newblock Spell: Self-play reinforcement learning for evolving long-context language models.
\newblock \emph{arXiv preprint arXiv:2509.23863}, 2025{\natexlab{b}}.

\bibitem[Yu et~al.(2023)Yu, Jiang, Shi, Yu, Liu, Zhang, Kwok, Li, Weller, and Liu]{yu2023metamath}
Longhui Yu, Weisen Jiang, Han Shi, Jincheng Yu, Zhengying Liu, Yu~Zhang, James~T Kwok, Zhenguo Li, Adrian Weller, and Weiyang Liu.
\newblock Metamath: Bootstrap your own mathematical questions for large language models.
\newblock \emph{arXiv preprint arXiv:2309.12284}, 2023.

\bibitem[Yu et~al.(2025)Yu, Lanchantin, Wang, Yuan, Golovneva, Kulikov, Sukhbaatar, Weston, and Xu]{yu2025cot}
Ping Yu, Jack Lanchantin, Tianlu Wang, Weizhe Yuan, Olga Golovneva, Ilia Kulikov, Sainbayar Sukhbaatar, Jason Weston, and Jing Xu.
\newblock Cot-self-instruct: Building high-quality synthetic prompts for reasoning and non-reasoning tasks.
\newblock \emph{arXiv preprint arXiv:2507.23751}, 2025.

\bibitem[Yuan et~al.(2024)Yuan, Pang, Cho, Li, Sukhbaatar, Xu, and Weston]{yuan2401self}
Weizhe Yuan, Richard~Yuanzhe Pang, Kyunghyun Cho, Xian Li, Sainbayar Sukhbaatar, Jing Xu, and Jason Weston.
\newblock Self-rewarding language models.
\newblock \emph{arXiv preprint arXiv:2401.10020}, 2024.

\bibitem[Yuan et~al.(2025)Yuan, Yu, Jiang, Padthe, Li, Kulikov, Cho, Wang, Tian, Weston, et~al.]{yuan2025naturalreasoning}
Weizhe Yuan, Jane Yu, Song Jiang, Karthik Padthe, Yang Li, Ilia Kulikov, Kyunghyun Cho, Dong Wang, Yuandong Tian, Jason~E Weston, et~al.
\newblock Naturalreasoning: Reasoning in the wild with 2.8 m challenging questions.
\newblock \emph{arXiv preprint arXiv:2502.13124}, 2025.

\bibitem[Yue et~al.(2024)Yue, Zheng, Zhang, and Chen]{yue2024mammoth2}
Xiang Yue, Tianyu Zheng, Ge~Zhang, and Wenhu Chen.
\newblock Mammoth2: Scaling instructions from the web.
\newblock \emph{Advances in Neural Information Processing Systems}, 37:\penalty0 90629--90660, 2024.

\bibitem[Zelikman et~al.(2022)Zelikman, Wu, Mu, and Goodman]{zelikman2022star}
Eric Zelikman, Yuhuai Wu, Jesse Mu, and Noah Goodman.
\newblock Star: Bootstrapping reasoning with reasoning.
\newblock In \emph{Advances in Neural Information Processing Systems}, 2022.

\bibitem[Zeng et~al.(2025)Zeng, Huang, Liu, Liu, He, Ma, and He]{simplerl}
Weihao Zeng, Yuzhen Huang, Qian Liu, Wei Liu, Keqing He, Zejun Ma, and Junxian He.
\newblock Simplerl-zoo: Investigating and taming zero reinforcement learning for open base models in the wild, 2025.
\newblock \url{https://arxiv.org/abs/2503.18892}.

\bibitem[Zhao et~al.(2025)Zhao, Wu, Yue, Wu, Xu, Lin, Wang, Wu, Zheng, and Huang]{zhao2025absolute}
Andrew Zhao, Yiran Wu, Yang Yue, Tong Wu, Quentin Xu, Matthieu Lin, Shenzhi Wang, Qingyun Wu, Zilong Zheng, and Gao Huang.
\newblock Absolute zero: Reinforced self-play reasoning with zero data.
\newblock \emph{arXiv preprint arXiv:2505.03335}, 2025.

\end{thebibliography}

\clearpage
\newpage
\beginappendix
\startcontents[sections]
\printcontents[sections]{l}{1}{\setcounter{tocdepth}{2}}
\newpage

\section{Evaluation Settings}
\label{app:eval_settings}

\textbf{Evaluation Protocol.} All models are evaluated in a zero-shot setting to assess whether the reasoning capabilities developed through corpus-grounded self-play transfer to standard benchmarks without task-specific adaptation. We use greedy decoding (temperature 0) for most evaluations to ensure reproducibility, following the evaluation protocols from~\cite{ma2025general}. For AIME'24 and AIME'25, we follow SimpleRL~\citep{simplerl} and report the average accuracy over 32 sampling runs with temperature 0.6 to better assess model performance on these challenging competition problems. For other mathematical reasoning benchmarks, we report pass@1 accuracy with greedy decoding for MATH-500~\citep{hendrycksmath2021}, OlympiadBench~\citep{he2024olympiadbench}, Minerva Math~\citep{lewkowycz2022solving}, GSM8K~\citep{cobbe2021gsm8k}, and AMC, evaluating exact match after answer extraction and normalization. All mathematical evaluations use GPT-4o~\citep{openai2024gpt4ocard} verification through the \texttt{simple-evals} framework, which handles equivalence checking for various mathematical formats including fractions, decimals, and algebraic expressions.

\textbf{General Reasoning Metrics.} For general reasoning benchmarks, we evaluate on GPQA-Diamond~\citep{rein2023gpqagraduatelevelgoogleproofqa}, the most challenging subset of graduate-level science questions designed to resist memorization; SuperGPQA~\citep{du2025supergpqa}, covering 285 disciplines with Google-search-resistant questions; MMLU-Pro~\citep{wang2024mmlupro}, an enhanced version of MMLU with more rigorous multiple-choice questions requiring deeper understanding; and BBEH~\citep{bbeh}, extending BIG-Bench Hard with additional complex reasoning tasks. All general reasoning evaluations use greedy decoding and exact match on the extracted answer choice (A/B/C/D) for multiple-choice questions. We maintain consistent prompting across all models, using the same system prompt and answer extraction format as during training to ensure fair comparison. The evaluation prompts instruct models to provide step-by-step reasoning before producing a final answer in a boxed format for mathematical tasks or selecting a letter choice for multiple-choice questions. All evaluation code and prompts will be released for reproducibility.

\section{Training Configuration Details}
\label{app:training_details}
\begin{table}[h!]
\centering
\small
\caption{Training configurations for all compared methods. All experiments use 640 iterations with batch size 128. $^\dagger$R-Zero shows performance degradation after 5 iterations, so we report its best performance. $^*$Absolute Zero applies -0.5 for incorrect but well-formatted responses and -1 for formatting errors.}
\label{tab:training_config}
\begin{tabular}{@{}lcccc@{}}
\toprule
\textbf{Configuration} & \textbf{SPICE} & \textbf{Strong Challenger} & \textbf{R-Zero} & \textbf{Absolute Zero} \\
\midrule
\multicolumn{5}{@{}l}{\textit{Data Source}} \\
Corpus documents & 20,000 & 20,000 & -- & -- \\
Question source & Document-grounded & Document-grounded & Self-generated & Self-generated \\
External grounding & \checkmark & \checkmark & $\times$ & Python executor \\
\midrule
\multicolumn{5}{@{}l}{\textit{Training Details}} \\
Challenger training & \checkmark & $\times$ (fixed) & \checkmark & \checkmark \\
Challenger sampling & 8 & -- & 4 & 1 \\
Reasoner training & \checkmark & \checkmark & \checkmark & \checkmark \\
Reasoner sampling & 8 & 8 & 5 & 1 \\
Temperature & 1.0 & 1.0 & 1.0 & 1.0 \\
Optimizer & DrGRPO & DrGRPO & GRPO & REINFORCE++ \\
\midrule
\multicolumn{5}{@{}l}{\textit{Reward Design}} \\
Challenger reward & Gaussian Variance (max 1.0) & N/A & $1-2|p-0.5|$ & $1-p$ \\
Reasoner reward & Binary correctness & Binary correctness & Binary correctness & Binary correctness \\
Invalid penalty & -0.1 & N/A & -1 (Challenger) & -0.5/-1$^*$ \\
\midrule
\multicolumn{5}{@{}l}{\textit{Performance}} \\
Training iterations & 640 & 640 & 5$^\dagger$ & 640 \\
\bottomrule
\end{tabular}
\end{table}

\textbf{Implementation Infrastructure.} All experiments utilize a distributed actor-learner architecture implemented in Oat~\citep{liu2024oat}, with vLLM~\citep{kwon2023efficient} for efficient inference during both question generation and answer sampling phases. We employ Math-Verify for answer equivalence checking, which handles various mathematical formats including fractions, decimals, and algebraic expressions. Training is conducted on 8 H200 GPUs per experiment with a learning rate of 1e-6 using constant learning rate schedule, gradient checkpointing, flash attention, and ZeRO Stage 2 optimization for memory efficiency. We use DrGRPO without KL regularization ($\beta = 0$) to focus purely on reward optimization. The system processes 128 trajectories per gradient update with rollout batch size of 128 distributed across devices. 

\textbf{Baseline Implementation Details.} R-Zero uses 5 rollouts for Reasoner training and 4 for Challenger training, with the Challenger computing uncertainty rewards $r_{\text{uncertainty}} = 1 - 2|p - 0.5|$ that peak when the Reasoner achieves 50\% accuracy, maximizing learning potential at the frontier of model capability~\citep{huang2025r}. Absolute Zero uses 1 rollout during training but estimates learnability from 8 Reasoner attempts, with rewards $r_{\text{propose}} = 1 - p$ (for $0 < p < 1$) that linearly decrease with solve rate and assign zero reward to trivial or impossible tasks. Both baselines use binary correctness rewards for their reasoners: R-Zero validates against majority-voted pseudo-labels using Math-Verify, while Absolute Zero checks Python value equality with ground-truth outputs. For invalid responses, R-Zero's Challenger assigns -1 penalty while its Reasoner uses weighted format+accuracy scoring; Absolute Zero applies -0.5 for incorrect but well-formatted responses and -1 for format errors. Our DrGRPO approach removes standard deviation normalization to avoid difficulty bias, computing $\hat{A}_i = R_i - \text{mean}(\{R_j\})$ to ensure gradient updates reflect genuine learning signal rather than question difficulty artifacts.

\textbf{Corpus Composition.} The 20,000 document corpus combines Nemotron-CC-Math~\citep{mahabadi2025nemotron} (50\%) for mathematical content and NaturalReasoning~\citep{yuan2025naturalreasoning} (50\%) for general reasoning, providing diverse contexts across STEM, humanities, and social sciences. Documents are sampled uniformly during training, with each document used approximately 2-3 times over 640 iterations. This diversity prevents overfitting to specific domains while ensuring broad coverage of reasoning patterns.

\section{Additional Results and Analysis}
\label{app:additional_results}

This appendix provides comprehensive benchmark-level results for all ablation studies presented in Section~\ref{sec:ablations}. These detailed breakdowns reveal how different design choices affect specific reasoning capabilities.

\subsection{Comprehensive Benchmark Results}\label{app:additional_benchmark_results}

We present complete results across all 11 evaluation benchmarks for each ablation configuration. Tables~\ref{tab:ablation_full}, \ref{tab:ablation_format_full}, and \ref{tab:ablation_reward_full} show the full performance breakdown, revealing nuanced patterns not visible in the averaged results.

\begin{table*}[htbp]
\centering
\small
\caption{Comprehensive ablation study on corpus composition effects (Qwen3-4B-Base). Best results per column in \textbf{bold}, second best \underline{underlined}.}
\label{tab:ablation_full}
\resizebox{\textwidth}{!}{%
\begin{tabular}{l|ccccccc|cccc|c}
\toprule
& \multicolumn{7}{c|}{\textbf{Mathematical Reasoning}} & \multicolumn{4}{c|}{\textbf{General Reasoning}} & \\
\cmidrule(lr){2-8} \cmidrule(lr){9-12}
\textbf{Corpus Type} & \textbf{AMC} & \textbf{Minerva} & \textbf{MATH500} & \textbf{GSM8K} & \textbf{Olympiad} & \textbf{AIME24} & \textbf{AIME25} & \textbf{SuperGPQA} & \textbf{GPQA-Dmd} & \textbf{MMLU-Pro} & \textbf{BBEH} & \textbf{Overall} \\
\midrule
DCLM (NaturalReasoning) & 47.5 & 46.8 & 71.6 & 86.3 & 37.2 & 9.8 & 11.9 & \textbf{31.8} & \textbf{42.4} & \textbf{60.3} & \textbf{13.5} & 41.7 \\
Math (Nemotron-CC-Math) & \textbf{62.5} & \textbf{56.8} & \textbf{81.2} & 91.8 & \textbf{46.3} & \textbf{14.8} & \textbf{22.1} & 26.2 & 30.8 & 52.7 & 9.6 & 43.2 \\
\rowcolor{red!10}
DCLM + Math (SPICE) & \underline{57.5} & \underline{51.9} & \underline{78.0} & \textbf{92.7} & \underline{42.7} & \underline{12.2} & \underline{19.1} & \underline{30.2} & \underline{39.4} & \underline{58.1} & \underline{12.3} & \textbf{44.9} \\
\bottomrule
\end{tabular}
}
\end{table*}

Several key insights emerge from the detailed corpus ablation (Table~\ref{tab:ablation_full}):
\begin{itemize}
    \item Mathematical benchmarks strongly favor Nemotron-CC-Math, with gains of up to +10.2 points on AIME25
    \item General reasoning benchmarks uniformly prefer NaturalReasoning, particularly GPQA-Diamond (+11.9) and BIG-Bench Extra Hard (+3.9)
    \item The combined approach achieves the highest GSM8K score (92.7\%), suggesting synergistic effects for certain problem types
\end{itemize}

\begin{table*}[htbp]
\centering
\small
\caption{Comprehensive ablation study on response format effects (Qwen3-4B-Base). Best results per column in \textbf{bold}, second best \underline{underlined}.}
\label{tab:ablation_format_full}
\resizebox{\textwidth}{!}{%
\begin{tabular}{l|ccccccc|cccc|c}
\toprule
& \multicolumn{7}{c|}{\textbf{Mathematical Reasoning}} & \multicolumn{4}{c|}{\textbf{General Reasoning}} & \\
\cmidrule(lr){2-8} \cmidrule(lr){9-12}
\textbf{Response Format} & \textbf{AMC} & \textbf{Minerva} & \textbf{MATH500} & \textbf{GSM8K} & \textbf{Olympiad} & \textbf{AIME24} & \textbf{AIME25} & \textbf{SuperGPQA} & \textbf{GPQA-Dmd} & \textbf{MMLU-Pro} & \textbf{BBEH} & \textbf{Overall} \\
\midrule
MCQ & \textbf{61.0} & 46.2 & 70.8 & 87.6 & 37.8 & 10.4 & 14.8 & \underline{30.6} & \textbf{41.8} & \textbf{59.4} & 12.1 & 42.0 \\
Free-form & 54.5 & \textbf{55.7} & \textbf{82.0} & 92.1 & \textbf{45.8} & \textbf{15.2} & \textbf{23.6} & 27.8 & 34.2 & 54.3 & 10.8 & 43.7 \\
\rowcolor{red!10}
MCQ + Free-form (SPICE) & \underline{57.5} & \underline{51.9} & \underline{78.0} & \textbf{92.7} & \underline{42.7} & \underline{12.2} & \underline{19.1} & \textbf{30.2} & \underline{39.4} & \underline{58.1} & \textbf{12.3} & \textbf{44.9} \\
\bottomrule
\end{tabular}
}
\end{table*}

The task type analysis (Table~\ref{tab:ablation_format_full}) reveals complementary strengths:
\begin{itemize}
    \item MCQs excel at AMC (61.0\%) where answer selection is the natural format
    \item Free-form dominates on complex mathematical problems requiring detailed solutions (MATH500: 82.0\%, AIME25: 23.6\%)
    \item The mixed approach achieves optimal GSM8K performance, suggesting that task diversity aids in word problem comprehension
\end{itemize}

The task type analysis (Table~\ref{tab:ablation_format_full}) reveals complementary strengths:
\begin{itemize}
    \item MCQs excel at AMC (61.0\%) where answer selection is the natural format
    \item Free-form dominates on complex mathematical problems requiring detailed solutions (MATH500: 82.0\%, AIME25: 23.6\%)
    \item The mixed approach achieves optimal GSM8K performance, suggesting that task diversity aids in word problem comprehension
\end{itemize}

\begin{table*}[htbp]
\centering
\small
\caption{Comprehensive ablation study on challenger reward strategies (Qwen3-4B-Base). Best results per column in \textbf{bold}, second best \underline{underlined}.}
\label{tab:ablation_reward_full}
\resizebox{\textwidth}{!}{%
\begin{tabular}{l|ccccccc|cccc|c}
\toprule
& \multicolumn{7}{c|}{\textbf{Mathematical Reasoning}} & \multicolumn{4}{c|}{\textbf{General Reasoning}} & \\
\cmidrule(lr){2-8} \cmidrule(lr){9-12}
\textbf{Challenger Reward} & \textbf{AMC} & \textbf{Minerva} & \textbf{MATH500} & \textbf{GSM8K} & \textbf{Olympiad} & \textbf{AIME24} & \textbf{AIME25} & \textbf{SuperGPQA} & \textbf{GPQA-Dmd} & \textbf{MMLU-Pro} & \textbf{BBEH} & \textbf{Overall} \\
\midrule
Absolute Zero Reward & 50.0 & 41.9 & 76.2 & 89.3 & 41.5 & 12.2 & 13.4 & 27.1 & 35.3 & 52.6 & 8.3 & 40.7 \\
Threshold & 52.5 & 43.7 & 75.8 & 89.7 & 40.8 & 11.8 & 14.2 & 27.8 & 36.2 & 53.4 & 9.1 & 41.4 \\
R-Zero Reward & \underline{55.0} & \underline{50.2} & \underline{77.4} & \underline{91.6} & \underline{41.8} & \underline{11.9} & 17.3 & \underline{29.4} & \underline{37.8} & \underline{56.3} & \underline{11.2} & \underline{43.6} \\
\rowcolor{red!10}
Variance (SPICE) & \textbf{57.5} & \textbf{51.9} & \textbf{78.0} & \textbf{92.7} & \textbf{42.7} & \textbf{12.2} & \textbf{19.1} & \textbf{30.2} & \textbf{39.4} & \textbf{58.1} & \textbf{12.3} & \textbf{44.9} \\
\bottomrule
\end{tabular}
}
\end{table*}

The reward strategy comparison (Table~\ref{tab:ablation_reward_full}) demonstrates the superiority of variance-based rewards:
\begin{itemize}
    \item Consistent improvements across all benchmarks compared to absolute zero baseline
    \item Largest gains on challenging benchmarks:  BIG-Bench Extra Hard (+4.0), AIME25 (+5.7), and MMLU-Pro (+5.5)
    \item The variance-based approach appears particularly effective for problems requiring multi-step reasoning
\end{itemize}

These detailed results confirm that \modelname{}'s design choices work synergistically to produce a well-rounded reasoning system, with each component contributing unique strengths that combine to achieve state-of-the-art performance.

\section{Challenger Reward Functions}
\label{app:reward_functions}

Figure~\ref{fig:reward_functions} visualizes the four challenger reward strategies compared in our ablation study. Each function represents a different philosophy for task selection during self-play:

\begin{figure}[h]
    \centering
    \includegraphics[width=0.5\linewidth]{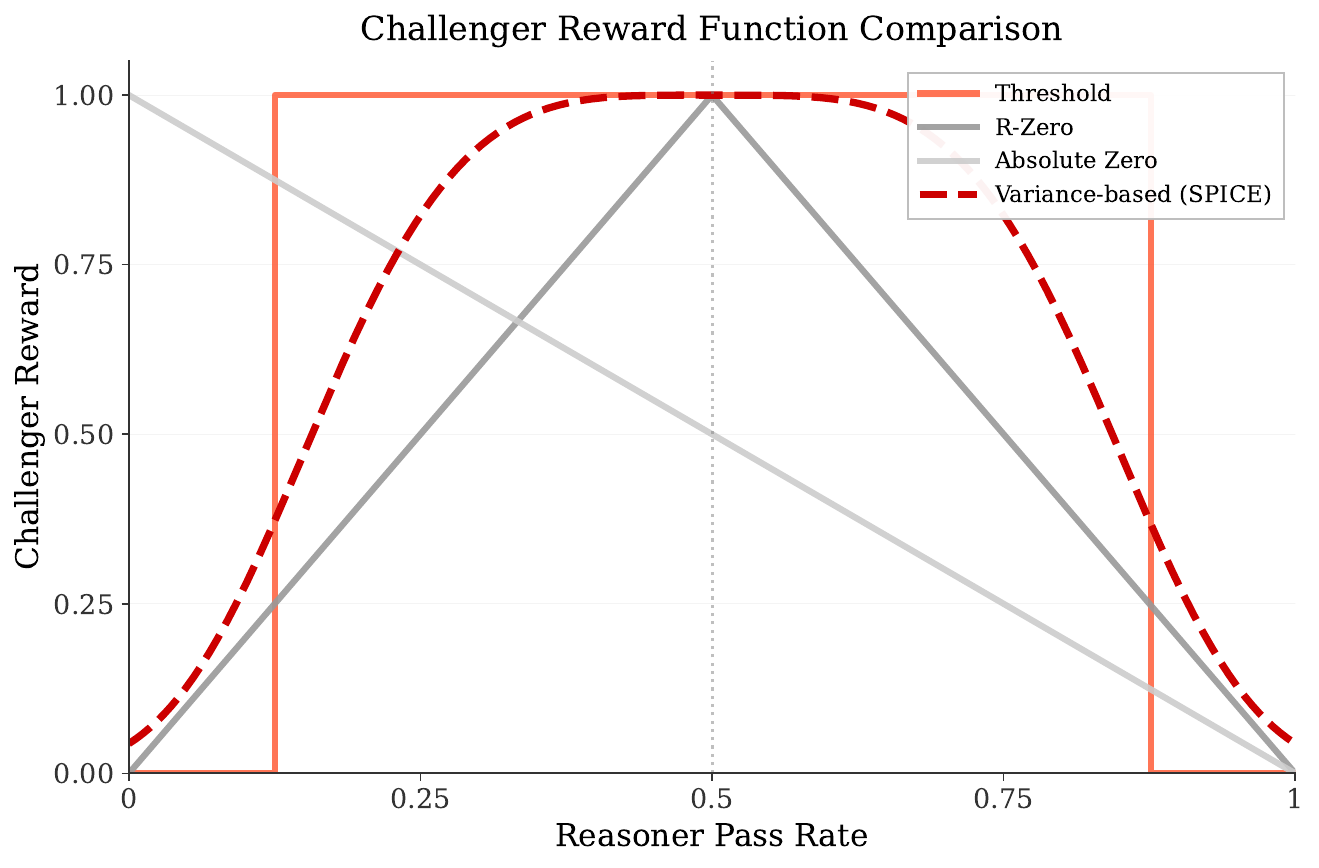}
    \caption{Comparison of challenger reward functions as a function of reasoner pass rate $p$. The \textbf{threshold reward} (rollout size 8) assigns 0 for trivial (8/8) and impossible (0/8) tasks, 1 otherwise. \textbf{Absolute Zero} ($1-p$) linearly decreases with pass rate. \textbf{R-Zero} ($1-2|p-0.5|$) peaks at maximum uncertainty. \textbf{\modelname{}'s variance-based reward} (dashed) uses $\exp(-\frac{(\sigma^2 - \sigma^2_{\text{opt}})^2}{2 \cdot 0.01})$ where $\sigma^2 = p(1-p)$ and $\sigma^2_{\text{opt}} = 0.25$, creating a smooth curriculum centered at the frontier of model capability.}
    \label{fig:reward_functions}
\end{figure}

The variance-based reward function used by \modelname{} is formulated as:
\begin{equation}
r_c = \exp\left(-\frac{(\sigma^2 - \sigma^2_{\text{opt}})^2}{2 \tau}\right)
\end{equation}
where $\sigma^2 = p(1-p)$ represents the variance of the reasoner's success rate, $\sigma^2_{\text{opt}} = 0.25$ is the optimal variance (achieved at $p=0.5$), and $\tau = 0.01$ is a temperature parameter controlling the sharpness of the reward peak.

This formulation ensures that:
\begin{itemize}
    \item Maximum reward occurs when the reasoner has a 50\% success rate, indicating problems at the edge of its capability.
    \item The reward smoothly decreases for both easier (high $p$) and harder (low $p$) problems.
    \item Unlike the threshold function, there are no discontinuities that could destabilize training.
    \item Unlike R-Zero, the Gaussian form provides better gradient signals near the optimum.
\end{itemize}

The empirical results in Table~\ref{tab:ablation_challenger_reward} demonstrate that this variance-based approach yields superior performance across all benchmarks, validating our hypothesis that optimal learning occurs when training focuses on problems with balanced difficulty.

\section{Prompt Templates}
\label{app:prompt_templates}

\textbf{Question Format Selection.} Before generating questions, the \C{} evaluates each document to determine the optimal question format using a structured prompt (Figure~\ref{fig:doc_eval_prompt}). This evaluation assesses whether the document contains complex relationships suitable for MCQs with plausible distractors, or precise information better suited for free-form questions with deterministic answers. The prompt returns a JSON response specifying the recommended format (MCQ or free-form) and, for free-form questions, the appropriate answer type (integer, expression, or string).

\textbf{Question Generation.} Following format selection, the \C{} uses specialized prompts to generate challenging questions (Figure~\ref{fig:task_gen_prompt}). The MCQ prompt guides generation through structured steps: identifying complex multi-concept relationships, creating balanced distractors, and self-testing difficulty—ensuring questions require synthesis rather than simple retrieval. The free-form prompt similarly emphasizes multi-step reasoning but focuses on generating questions with unambiguous, extractable answers. Both prompts enforce that questions must be self-contained and solvable without the source document, preventing reference-dependent questions that would be unanswerable for the \R{}.

\textbf{Answer Generation.} The \R{} uses model-specific templates optimized for each architecture. Qwen3 models use chat-style formatting with explicit instructions to reason step-by-step before providing boxed answers, while OctoThinker models employ a conversational format emphasizing internal reasoning processes. These templates ensure consistent answer extraction across different model families while maintaining their optimal performance characteristics.

\begin{figure}[h!]
\centering
\footnotesize
\begin{AIbox}{Prompt Templates for Different Model Families}
\textbf{Qwen3 Family Template:}
\begin{quote}
\texttt{<|im\_start|>system}\\
\texttt{You are a helpful assistant.}\\
\texttt{<|im\_end|>}\\
\texttt{<|im\_start|>user}\\
\texttt{\{question\}}\\
\\
\texttt{Please reason step by step, and put your final answer within \textbackslash boxed\{\}.}\\
\texttt{<|im\_end|>}\\
\texttt{<|im\_start|>assistant}
\end{quote}
\vspace{1em}
\textbf{OctoThinker Family Template:}
\begin{quote}
\texttt{A conversation between User and Assistant. The user asks a question, and the Assistant solves it. The assistant first thinks about the reasoning process in the mind and then provides the user with the answer.}\\
\\
\texttt{User: \{question\}}\\
\texttt{You must put your answer inside \textbackslash boxed\{\}.}\\
\\
\texttt{Assistant:}
\end{quote}
\end{AIbox}
\caption{Model-specific prompt templates for the \R{} role. The \C{} role uses similar formatting with task-specific instructions for question generation from documents. The \texttt{\{question\}} placeholder is replaced with the actual question during inference.}
\label{fig:prompt_templates}
\end{figure}

\begin{figure}[h!]
\centering
\begin{tcolorbox}[
    colframe=mediumred,
    colback=verylightred,
    title=Task Type Selection Prompt,
    fonttitle=\bfseries\small
]
\fontsize{6.7pt}{7pt}\selectfont
\texttt{Analyze this document and decide whether it's better suited for a CHALLENGING multiple-choice question (MCQ) or a free-form question.}\\
\texttt{Document:}\\
\texttt{\{document\}}\\
\\
\texttt{Consider the prompts that will be used:}\\
\\
\texttt{For MCQ:}\\
\texttt{- Needs complex relationships and multi-step reasoning paths}\\
\texttt{- Should allow creating 3 plausible but wrong distractors}\\
\texttt{- Requires synthesis of multiple concepts}\\
\texttt{- Can test understanding through carefully crafted wrong answers}\\
\\
\texttt{For Free-form:}\\
\texttt{- Best for questions requiring specific calculations (Integer answers)}\\
\texttt{- Good for deriving formulas or expressions (Expression answers)}\\
\texttt{- Suitable for conceptual answers requiring precise terminology (String answers)}\\
\texttt{- Should have a single clear correct answer}\\
\\
\texttt{Based on the document content, choose EXACTLY ONE type that would produce the highest quality CHALLENGING question.}\\
\\
\texttt{You MUST respond with ONLY a valid JSON object (no markdown, no explanation before or after):}\\
\texttt{\{\{}\\
\texttt{  "suitable\_for\_mcq": <true or false>,}\\
\texttt{  "suitable\_for\_free\_form": <true or false>,}\\
\texttt{  "best\_answer\_type": <"Integer" or "Expression" or "String" or null>,}\\
\texttt{  "reason": "<brief explanation without special characters>"}\\
\texttt{\}\}}\\
\\
\texttt{CRITICAL RULES:}\\
\texttt{1. Return ONLY the JSON object, no other text}\\
\texttt{2. Exactly ONE of suitable\_for\_mcq or suitable\_for\_free\_form must be true}\\
\texttt{3. Do NOT use backticks or markdown formatting}\\
\texttt{4. Do NOT include LaTeX or special characters in the reason field}\\
\texttt{5. Keep reason under 100 characters}

\end{tcolorbox}
\caption{Task type selection prompt used by the \C{} to determine optimal question format. The prompt analyzes document content characteristics and returns a structured JSON response specifying whether the document better supports MCQ generation (with plausible distractors) or free-form questions (with deterministic answers). This format selection ensures questions leverage the document's content effectively.}
\label{fig:doc_eval_prompt}
\end{figure}

\begin{figure}[h]
\centering
\begin{tcolorbox}[
    colframe=mediumred,
    colback=verylightred,
    title=MCQ Challenger Prompt,
    fonttitle=\bfseries\small
]
\fontsize{6pt}{7pt}\selectfont
Your task is to create CHALLENGING exam questions from a document by identifying complex relationships and multi-step reasoning paths.

\#\# Text
[BEGINNING OF THE DOCUMENT]
\{text\}
[END OF THE DOCUMENT]
\\
\\
\#\# Instructions
\\
\\
\#\#\# Step 1: Complex Information Extraction
**PRIORITY: Focus on information that requires synthesis and reasoning**
Scan the text and identify information that requires connecting multiple concepts:
* Relationships between multiple variables or concepts that span different sections
* Multi-step calculations or procedures where each step depends on previous ones
* Formulas or principles that require understanding interactions between components
* Implicit conclusions that can be derived by combining stated facts
* Comparative analyses or trade-offs between different approaches
* Conditional relationships (if X then Y, but if Z then W)
* Systems where changing one parameter affects multiple others
\\
\\
**AVOID**: 
* Single, directly stated facts (these create Easy questions)
* Simple definitions that stand alone
* Values or numbers mentioned in isolation
* Information that requires no synthesis
\\
\\
\#\#\# Step 2: Difficulty Enhancement Process
**EXPLICITLY STATE YOUR HARDENING PROCESS**
Before generating the question, describe your strategy to make it harder:
1. What simple version would you avoid?
2. What complexity layers will you add?
3. Which concepts will you force students to connect?
4. What common shortcuts will you block?
5. How will you ensure multi-step reasoning is required?
\\
\\
Document this in the output field `"hardening\_process"`.
\\
\\
\#\#\# Step 3: Advanced Question Generation
For each complex relationship identified, create a question that:
* Requires applying multiple concepts from different parts of the document
* Tests understanding of relationships, not just recall of facts
* Forces reasoning through multiple steps to reach the answer
* May require comparing or contrasting different scenarios
* Could involve "what if" scenarios based on principles in the text
* Tests ability to apply concepts to slightly modified situations
\\
\\
**CRITICAL - Self-Contained Requirements**:
* Questions must be 100\% self-contained and standalone
* NEVER use: "according to the text", "in the document", "as mentioned", "the passage states", "based on the analysis", etc.
* Write as if for a formal exam with no reference material
* Include all necessary context within the question itself
* Define any specialized terms if needed for clarity
\\
\\
\#\#\# Step 4: Difficulty-Driven Design
**TARGET: Generate HARD/EXTRA HARD questions by design**
* HARD: Synthesize 4+ concepts; multi-step problem solving; pattern recognition
* EXTRA HARD: Complex system analysis; counter-intuitive applications; edge cases
\\
\\
Design questions that CANNOT be answered by:
* Looking up a single fact
* Finding one sentence with the answer
* Simple keyword matching
\\
\\
\#\#\# Step 5: Knowledge Integration Requirements
Document the reasoning path that shows why this is a difficult question:
* List 3+ distinct pieces of information needed from different parts
* Show the logical connections required between these pieces
* Explain why simple lookup won't work
* Include intermediate reasoning steps
\\
\\
\#\#\# Step 6: Multiple Choice Design Guidelines
Create a multiple choice question with 4 options following these STRICT rules:
\\
\\
**Length Balance**: All options must be approximately equal length (±20%
**Unit Consistency**: All numerical answers must use identical units and formatting
**Tone Neutrality**: Avoid overly certain language ("definitely", "always", "never") unless justified
**Plausibility**: All distractors must be genuinely plausible based on partial understanding
\\
\\
Format:
Question: [Complete, self-contained question with all necessary context]
A) [Balanced length option]
B) [Balanced length option]
C) [Balanced length option]
D) [Balanced length option]
Correct: [Letter]
\\
\\
**Distractor Design**:
* Common calculation errors from the multi-step process
* Results from applying only partial reasoning
* Mixing up related concepts from the document
* Reasonable approximations that miss key factors
\\
\\
\#\#\# Step 7: Self-Testing Filter (AFTER MCQ Creation)
**SOLVE YOUR OWN MCQ AS A STUDENT WOULD**
Now test the complete multiple choice question:
1. What's the quickest path a student might try with these options?
2. Can you eliminate 2+ options without full understanding? If yes, redesign distractors
3. Does seeing the options make the answer obvious? If yes, improve distractors
4. Count the reasoning steps required even with options visible - if less than 3, REJECT
5. Time estimate: Would this MCQ take <30 seconds? If yes, make it harder
6. Could a student guess correctly by pattern matching the options? If yes, rebalance
\\
\\
Document your solving process in `"self\_test\_solution"`.
\\
\\
\#\#\# Step 8: Final Complexity Verification
Before finalizing, verify your question is NOT Easy by checking:
* Can it be answered by finding one sentence? If yes, redesign
* Does it require connecting multiple document sections? If no, add complexity
* Would someone need to understand relationships, not just facts? If no, refocus
* Are all MCQ options balanced and using consistent formatting? If no, revise
* Did your self-test of the MCQ take more than 1 minute? If no, increase difficulty
\\
\\
\#\# Output Format
FIRST, think step-by-step about your question design (this is your private thinking).
\\
\\
THEN, provide your complete analysis in a JSON object with these fields.
CRITICAL: Output ONLY valid JSON without any markdown formatting or code blocks.
DO NOT wrap your JSON in ```json``` or any other markers.
Start directly with {{ and end with }}
\\
\\
Example CORRECT format (copy this structure):
\{\{"identified\_answer": "your answer", "answer\_quote": ["quote1", "quote2"], "hardening\_process": "strategy"\}\}
\\
\\
Example WRONG format (DO NOT do this):
```json
{{"identified\_answer": "your answer"}}
```
\\
\\
- `"identified\_answer"`: The complex relationship or multi-step conclusion derived from synthesizing document content

- `"answer\_quote"`: Multiple relevant quotes showing the different pieces needed (not just one quote)

- `"hardening\_process"`: Your explicit strategy for making this question difficult (from Step 2)

- `"exam\_question"`: A challenging, self-contained question requiring synthesis. Return empty string if document lacks sufficient complexity.

- `"correct\_answer"`: Complete answer showing the reasoning chain using document content. Return empty string if not derivable from document.

- `"multiple\_choice\_question"`: Self-contained MC question with balanced options. Return empty string if no question generated.

- `"multiple\_choice\_correct"`: The correct option letter (A, B, C, or D). Return empty string if no MC question.

- `"self\_test\_solution"`: Your step-by-step solution of the MCQ showing the difficulty (from Step 7)

- `"knowledge\_and\_reasoning\_steps"`: Detailed reasoning path showing why this is Hard/Extra Hard difficulty.

- `"question\_difficulty"`: Target difficulty (Hard/Extra Hard). Avoid "Easy" unless document truly lacks complexity.

\end{tcolorbox}
\label{fig:challenger_mcq_prompt}
\end{figure}

\begin{figure}[h]
\centering
\begin{tcolorbox}[
    colframe=mediumred,
    colback=verylightred,
    title=Free-form Question Challenger Prompt,
    fonttitle=\bfseries\small
]
\fontsize{6pt}{7pt}\selectfont
Your task is to create CHALLENGING free-form questions from a document that require deep understanding and complex reasoning.
\\
\\
\#\# Text
[BEGINNING OF THE DOCUMENT]
\{text\}
[END OF THE DOCUMENT]
\\
\\
\#\# Answer Type
You must generate a question with answer type: {answer\_type}
\\
\\
\#\# Instructions
\\
\\
\#\#\# Step 1: Complex Information Extraction for {answer\_type}
**PRIORITY: Focus on information that requires synthesis and multi-step reasoning**
Based on the answer type {answer\_type}, scan the text and identify:
\\
\\
**For Integer/Float answers:**
* Multi-variable calculations spanning different sections
* Sequential computations where each step depends on previous results
* Counting problems requiring careful categorization
* Rate/ratio/percentage problems with multiple components
* Optimization problems with constraints
* Statistical calculations requiring data aggregation
\\
\\
**For Expression answers:**
* Relationships between multiple variables that form equations
* Patterns that can be generalized into formulas
* Systems of equations from different constraints
* Derivative relationships or functional dependencies
* Algebraic expressions combining multiple principles
* Recursive or iterative formulas
\\
\\
**For String answers (MUST BE CONCISE):**
* Single words or short phrases (1-3 words maximum)
* Technical terms, names, or identifiers
* Categories or classifications (single term only)
* Named entities (person, place, concept, method name)
* Units, symbols, or abbreviated forms
* AVOID: Long descriptions, sentences, or explanations
* Examples: "Newton", "TCP/IP", "gradient descent", "Paris", "O(n log n)"
\\
\\
**For Boolean answers:**
* Complex logical conditions with multiple clauses
* Statements requiring verification across multiple facts
* Comparative claims needing multi-point analysis
* Existence or uniqueness proofs
* Conditional truths depending on context
* Negations requiring comprehensive checking
\\
\\
**AVOID**:
* Direct lookups or single-fact answers
* Simple arithmetic or basic calculations
* Definitions stated verbatim in text
* Trivial yes/no questions
\\
\\
\#\#\# Step 2: Difficulty Enhancement Strategy
**EXPLICITLY STATE YOUR HARDENING PROCESS**
Before generating the question, document your strategy:
1. What simple version would be too easy?
2. What complexity layers will you add?
3. Which document sections must be synthesized?
4. What intermediate steps are required?
5. How will you prevent shortcut solutions?
6. What makes this require deep understanding?
\\
\\
Document this in `"hardening\_process"`.
\\
\\
\#\#\# Step 3: Advanced Question Generation for {answer\_type}
Create a question that:
* Requires connecting 3+ concepts from different parts
* Cannot be answered by simple lookup or keyword matching
* Forces multi-step reasoning to reach the answer
* Tests understanding of relationships, not memorization
* May involve applying principles to modified scenarios
* Requires precise interpretation for the specific answer type
\\
\\
**CRITICAL - Self-Contained Requirements**:
* Questions must be 100%
* NEVER use: "according to the text", "in the document", "as mentioned", etc.
* Write as if for a formal exam with no reference material
* Include all necessary context and definitions
* Specify units, formats, or constraints clearly
\\
\\
\#\#\# Step 4: Answer Precision for {answer\_type}
**CRITICAL - Answer Format Requirements**:
\\
\\
**Integer answers:**
* Must be whole numbers only
* Specify units if applicable (e.g., "in meters", "number of items")
* No decimals, fractions, or ranges
* CORRECT JSON: "answer": 42 or "answer": "42"
* WRONG JSON: "answer": [42] or "answer": {{"value": 42}}
\\
\\
**Float answers:**
* Specify precision required (e.g., "to 2 decimal places")
* Include units if applicable
* Use decimal notation, not fractions
* Example: 3.14, not "$\pi$" or "$22/7$"
\\
\\
**Expression answers:**
* Use standard mathematical notation
* Variables must be clearly defined
* Simplify to canonical form
* CORRECT JSON: "answer": "2*x\^2 + 3*x - 5"
* WRONG JSON: "answer": ["2*x\^2 + 3*x - 5"] or "answer": {{"expr": "2*x\^2"}}
\\
\\
**String answers:**
* Specify exact format expected
* Case sensitivity requirements
* No extra punctuation or quotes
* CORRECT JSON: "answer": "Newton's Third Law"
* WRONG JSON: "answer": ["Newton's Third Law"] or "answer": {{"text": "Newton"}}
\\
\\
**List answers:**
* Specify ordering (alphabetical, chronological, by magnitude)
* Delimiter format (comma-separated, JSON array)
* Whether duplicates are allowed
* Example: ["apple", "banana", "cherry"] for JSON format
\\
\\
**Boolean answers:**
* Must be exactly "true" or "false" (lowercase)
* No "yes/no", "T/F", or other variations
* Clear truth conditions
* Example: true, not "True" or "yes"
\\
\\
\#\#\# Step 5: Solution Verification Process
**SOLVE YOUR OWN QUESTION STEP-BY-STEP**
Work through the complete solution:
1. Identify all required information pieces
2. Show each calculation or reasoning step
3. Handle any edge cases or special conditions
4. Arrive at the final answer in the correct format
5. Verify the answer matches the specified type exactly
6. Estimate solving time (should be >1 minute for hard questions)
\\
\\
Document your solution in `"step\_by\_step\_solution"`.
\\
\\
\#\#\# Step 6: Difficulty Calibration
Rate your question's difficulty and justify:
\\
\\
**MEDIUM (2-3 steps, 1-2 minutes):**
* Requires combining 2-3 document sections
* Clear path once relationships identified
* Some calculation or reasoning required
\\
\\
**HARD (4-5 steps, 2-3 minutes):**
* Synthesizes 4+ concepts
* Multiple valid approaches possible
* Requires careful analysis to avoid errors
\\
\\
**EXTRA HARD (6+ steps, 3+ minutes):**
* Complex system with many interactions
* Counter-intuitive results possible
* Requires deep understanding of principles
\\
\\
Document reasoning in `"difficulty\_justification"`.
...

\end{tcolorbox}
\end{figure}

\begin{figure}[tp]
\centering
\begin{tcolorbox}[
    colframe=mediumred,
    colback=verylightred,
    title=Free-form Question Challenger Prompt (continued),
    fonttitle=\bfseries\small
]
\fontsize{6pt}{7pt}\selectfont
\#\#\# Step 7: Alternative Interpretations Check
**ENSURE UNAMBIGUOUS ANSWER**
Verify your question has exactly ONE correct answer:
1. Could the question be interpreted differently?
2. Are all constraints clearly specified?
3. Is the answer format unambiguous?
4. Would different valid approaches yield the same answer?
5. Are edge cases properly handled?
\\
\\
If ambiguous, revise the question for clarity.
\\
\\
\#\#\# Step 8: Final Complexity Verification
Before finalizing, verify:
* Cannot be answered by simple text search
* Requires understanding, not just extraction
* Answer type matches {answer\_type} exactly
* Solution requires multiple reasoning steps
* Question is self-contained and clear
* Difficulty matches your target level
\\
\\
\#\# CRITICAL ANSWER FORMAT RULES - MUST FOLLOW EXACTLY
\\
\\
The "answer" field MUST be a simple value, NOT nested in lists or dicts.
The "question" field MUST be a single string, NOT a list of questions.
\\
\\
CORRECT formats:
- question: "What is 2+2?" (NOT ["What is 2+2?"] or ["Q1", "Q2"])  
- Integer answer: 42 or "42" (NOT [42] or {{"value": 42}})
- Expression answer: "2*x + 5" (NOT ["2*x + 5"] or {{"expr": "2*x + 5"}})
- String answer: "Paris" or "TCP/IP" (1-3 words max, NOT ["Paris"] or {{"answer": "Paris"}})
\\
\\
WRONG formats that will FAIL:
- question: ["What is 2+2?"] - Don't return list of questions!
- question: ["Q1: ...", "Q2: ..."] - Return ONLY ONE question!
- answer: [42] - Don't wrap in list!
- answer: {{"value": 42}} - Don't wrap in dict!
- answer: {{"answer": "Paris"}} - Don't nest the answer!
\\
\\
\#\# Output Format
FIRST, think step-by-step about your question design (this is your private thinking).
\\
\\
THEN, provide your complete analysis in a JSON object with these fields.
CRITICAL: Output ONLY valid JSON without any markdown formatting or code blocks.
DO NOT wrap your JSON in ```json``` or any other markers.
Start directly with {{ and end with }}
\\
\\
Example CORRECT format (copy this structure):
{{"identified\_information": ["info1", "info2"], "question": "your question", "answer": 42, "answer\_type": "Integer"}}
\\
\\
Example WRONG format (DO NOT do this):
```json
{{"question": "your question", "answer": 42}}
```
\\
\\
- `"identified\_information"`: List the 3+ key pieces of information from different document sections needed to solve
- `"relevant\_quotes"`: Include multiple verbatim quotes from the document showing the different pieces needed
- `"hardening\_process"`: Describe your explicit 4-step strategy for making this question difficult
- `"question"`: EXACTLY ONE complete, challenging, self-contained question as a single string (NOT a list, NOT multiple questions). Return empty string if document lacks complexity for {answer\_type} questions.
- `"answer"`: The precise answer in correct {answer\_type} format. Return empty string if not derivable from document.
- `"answer\_type"`: Must be exactly "{answer\_type}"
- `"step\_by\_step\_solution"`: List each step of the complete solution showing the reasoning chain
- `"intermediate\_results"`: Dictionary of intermediate calculations or conclusions from each step
- `"difficulty\_level"`: Either "Hard" or "Extra Hard" (no Medium for this complexity level)
- `"difficulty\_justification"`: Explain why this specific difficulty rating based on steps and concepts required
- `"solving\_time\_estimate"`: Realistic estimate in minutes for a student to solve
- `"required\_concepts"`: List the specific concepts from the document that must be understood
- `"potential\_errors"`: Common mistakes or edge cases students might encounter
\\
\\
**CRITICAL RULES**:
1. If document lacks complexity for {answer\_type}, return {{"question": "", "answer": "", "answer\_type": "{answer\_type}"}}
2. Answer field must EXACTLY match {answer\_type} format requirements
3. Never reference "the document" or "the text" in the question
4. Ensure answer is derivable from provided document content
5. Question must be solvable with document information alone
6. For String answers: MAXIMUM 3 words - prefer single terms, names, or short identifiers
\end{tcolorbox}
\caption{\C{} Prompts. Depending on the document, the \C{} uses either the MCQ or Free-form prompt to generate tasks for the \R{}.}
\vspace{500pt}
\label{fig:task_gen_prompt}
\end{figure}

\end{document}